\documentclass[a4paper,fleqn]{cas-dc}

\usepackage[authoryear,longnamesfirst]{natbib}
\usepackage{threeparttable}

\def\tsc#1{\csdef{#1}{\textsc{\lowercase{#1}}\xspace}}
\tsc{WGM}
\tsc{QE}

\usepackage{enumitem}
\usepackage{graphicx}

\begin{document}
\let\WriteBookmarks\relax
\def\floatpagepagefraction{1}
\def\textpagefraction{.001}

\shorttitle{Particle Competition and Cooperation for Robust Graph Convolutional Network Learning Under Label Noise}    

\shortauthors{F. Breve}  

\title [mode = title]{Particle Competition and Cooperation for Robust Graph Convolutional Network Learning Under Label Noise}  

\tnotemark[1] 

\tnotetext[1]{} 

%

\author[1]{Fabricio Breve}[orcid=0000-0002-1123-9784]
\cormark[1]
\ead{fabricio.breve@unesp.br}
\affiliation[1]{organization={São Paulo State University - UNESP},
	city={Rio Claro - SP},
	country={Brazil}}


\begin{abstract}
	Graph Convolutional Networks (GCNs) are highly sensitive to label noise, since corrupted supervision can propagate through the graph and degrade learned node representations. This work proposes PCC+GCN, a hybrid framework that uses Particle Competition and Cooperation (PCC) as a graph-based label-refinement stage before GCN training. PCC identifies suspicious labeled nodes through particle domination dynamics 	and determines whether their labels should be preserved, removed, or reassigned before GCN training. The framework also allows the graph used by PCC to be augmented with feature-based $k$-nearest-neighbor edges, while the GCN itself is trained on the original graph structure and node features. The proposed method was evaluated on ten graph datasets from the NoisyGL benchmark under conventional Uniform, Pair, and Random label noise, as well as under instance-dependent label noise. A detailed hyperparameter analysis was also conducted on Cora, CiteSeer, and PubMed. Under conventional noise, PCC+GCN achieved the highest overall average accuracy and the best average rank among the evaluated methods, with an average gain of $1.67$ percentage points over the baseline GCN across the clean setting and all noisy scenarios. Under instance-dependent noise, PCC+GCN remained competitive with the best-performing robust methods while requiring substantially lower execution time, being the fastest robust method on eight of the ten datasets. The results indicate that PCC-based label refinement provides an effective and computationally efficient preprocessing strategy for improving GCN robustness under noisy supervision.
\end{abstract}


\begin{highlights}
	\item PCC refines noisy labels before GCN training without modifying the GCN
	\item PCC+GCN improves robustness across multiple conventional label-noise models
	\item PCC+GCN achieves the best overall average rank in the NoisyGL benchmark
	\item PCC+GCN remains competitive under instance-dependent label noise
	\item PCC+GCN is the fastest robust method on eight of ten datasets
\end{highlights}



\maketitle

\section{Introduction}\label{Introduction}

Graph-structured learning has become a central paradigm for modeling data in which the relationships among instances are as informative as the instances themselves. In this context, Graph Neural Networks (GNNs) \citep{scarselli2009graph} have achieved strong performance in semi-supervised node classification and graph classification tasks because they exploit relational structure through message passing and neighborhood aggregation. Among these models, Graph Convolutional Networks (GCNs) \citep{kipf2017semi} are particularly influential because they generalize convolution-like operations to non-Euclidean domains and provide a simple yet effective framework for learning from graph-structured data. However, despite their success, GCNs remain highly sensitive to noisy supervision, since mislabeled nodes can propagate incorrect information through the graph and degrade the learned representations; moreover, this issue is common in real-world graph data \citep{wang2024noisygl,dai2021graph,xia2021towards}.

Particle Competition and Cooperation (PCC) is a graph-based semi-supervised learning model that has been applied to a variety of problems, including classification \citep{breve2012particle}, fuzzy community detection \citep{breve2013fuzzy}, image segmentation \citep{breve2015interactive}, image retrieval \citep{kawai2025semi}, and outlier repositioning \citep{pereira2026clustering}. In PCC, particles associated with different classes compete to dominate nodes, while particles from the same class cooperate, yielding a dynamic mechanism well suited to label refinement and propagation. A PCC variant specifically designed for label noise \citep{breve2015particle} further demonstrated that the method can classify unlabeled nodes and relabel noisy ones in the same process, providing a graph-aware denoising mechanism without a separate correction stage. These properties make PCC attractive when labels are scarce or unreliable.

At the same time, PCC has an important limitation in graph learning applications: it relies on a graph that must be provided or constructed beforehand, typically from the dataset’s native relational structure or from a k-nearest-neighbor graph in feature space. As a result, its performance is strongly affected by graph quality and graph construction choices, especially in settings where the underlying topology is incomplete or only partially observed. In this sense, PCC is naturally suited for label refinement, whereas GCNs provide stronger representation learning capabilities for downstream node classification.

These observations motivate a hybrid strategy that combines PCC with GCNs. Rather than using PCC as a standalone classifier, we propose to use it as a denoising stage that performs confidence-aware label refinement before GCN training. The GCN then performs the final classification over the graph structure, benefiting from cleaner supervision while still leveraging the expressive power of GCNs. The central hypothesis is that PCC and GCNs are complementary: PCC provides a graph-aware label correction mechanism that reduces the impact of noisy labels early in the pipeline, and the GCN leverages the cleaned supervision to learn more reliable embeddings and improve classification performance under label noise.

The main contributions of this work are threefold:
\begin{enumerate}[label=(\roman*)]
	\item a hybrid PCC+GCN framework for graph learning under noisy labels;
	\item an extensive hyperparameter sensitivity analysis for the proposed PCC+GCN framework; and
	\item an empirical comparison against representative robust	graph learning baselines under conventional and instance-dependent label-noise models, including an evaluation of computational efficiency.
\end{enumerate}

To evaluate this hybrid strategy, the remainder of the paper presents three sets of experiments. First, we conduct a hyperparameter study to identify the most sensitive parameters and their best ranges for the proposed PCC+GCN pipeline. Second, we compare the method against representative label-noise-robust graph learning methods from the NoisyGL benchmark, including NRGNN \citep{dai2021nrgnn}, PIGNN \citep{du2023noiserobust}, and CP \citep{zhang2020adversarial}, under uniform, random, and pairwise noise. Third, we test the approach under instance-dependent label noise to assess whether the proposed denoising strategy remains effective when corruption is correlated with the input structure, while also evaluating classification accuracy and computational efficiency.

The remainder of the paper is organized as follows. Section~\ref{RelatedWork} reviews related work on robust graph learning under noisy labels, PCC-based label refinement, and instance-dependent label noise. Section~\ref{sec:ProposedFramework} presents the proposed PCC+GCN framework, while Section~\ref{sec:ExperimentalSetup} describes the experimental setup. Section~\ref{sec:HyperparameterAnalysis} analyzes the sensitivity of the PCC hyperparameters. Section~\ref{sec:BenchmarkComparison} compares the proposed method with the methods included in the NoisyGL benchmark under conventional label-noise models. Section~\ref{sec:InstanceNoise} evaluates the methods under instance-dependent label noise, including both classification performance and computational efficiency. Finally, Sections~\ref{sec:Discussion} and~\ref{sec:Conclusion} present the discussion and conclusions, respectively.

\section{Related Works}\label{RelatedWork}

This section reviews related work on robust graph learning under noisy labels, graph-based label refinement, and instance-dependent label noise. We first discuss recent methods designed to improve the robustness of graph neural networks to corrupted labels, followed by previous work on Particle Competition and Cooperation (PCC) and its applications to graph-based semi-supervised learning and label refinement. Finally, we discuss recent work on instance-dependent noise and its relevance for evaluating robust graph learning methods under more realistic corruption processes.

\subsection{Graph Neural Networks under Label Noise}

Graph Neural Networks (GNNs) have achieved strong results in semi-supervised node classification, but their dependence on message passing makes them especially vulnerable to label noise, since corrupted labels can spread through the graph and affect neighboring representations \citep{wang2024noisygl,chiu2026hrgnn}. Consequently, robust graph learning under label noise remains an open challenge, motivating methods designed to reduce the impact of noisy labels during training.

A number of robust graph-learning methods have been proposed to address label noise from different perspectives. Some approaches focus on label correction or sample selection. For instance, \citet{wu2024robust} address noisy labels in heterophilic anomaly-detection graphs by augmenting graph structure and refining labels through small-loss-based node selection and confidence-aware training. Similarly, \citet{du2023noiserobust} introduce PIGNN, which improves robustness to label noise by estimating pairwise node interactions and using them to regularize node classification through decoupled training. In the same line, \citet{zhang2025efficient} propose Adaptive Label Refinement, which adaptively converts hard labels into soft labels and progressively sharpens confident samples through an entropy-based objective. Earlier work by \citet{li2021unified} proposes a unified robust training framework that performs label aggregation, sample reweighting, and label correction end to end for noisy graph node classification.

Other methods improve robustness through modified training objectives and noise governance. \citet{qian2023robust} propose RTGNN, which explicitly distinguishes clean and noisy labels and combines self-reinforcement with consistency regularization to reduce overfitting to corrupted supervision. In a related direction, \citet{jin2025systematic} systematically study the failure modes of GNNs under label noise and show that different architectures exhibit markedly different levels of vulnerability. Together, these works suggest that robust graph learning can benefit from both robust training objectives and a better understanding of how label noise affects optimization and generalization.

A third line of work exploits graph topology to improve sample selection or label refinement. \citet{wu2024mitigating} introduce Topological Sample Selection, which uses topological information to select informative nodes under noisy supervision. In a related direction, \citet{wang2025learning} propose Topological Feature Reconstruction, which leverages clean-label patterns to reconstruct graph features by exploiting topological information and improve robustness to label noise. These methods highlight that graph structure itself can be exploited to identify cleaner and more informative samples during training.

Although these approaches improve robustness through topology-aware selection, adaptive training, or label refinement, most methods remain tightly coupled to end-to-end GNN optimization. In contrast, our work investigates whether a graph-based label refinement mechanism based on PCC can serve as a lightweight denoising stage prior to GCN training.

\subsection{Particle Competition and Cooperation and Graph-Based Label Refinement}

Particle Competition and Cooperation (PCC) was originally proposed as a graph-based semi-supervised learning method for label propagation in networks \citep{breve2012particle}. In this framework, labeled nodes act as particle home nodes, and particles belonging to the same class cooperate while competing with particles from other classes, allowing labels to spread according to local graph structure and domination dynamics. This design makes PCC particularly suitable for semi-supervised learning scenarios in which neighboring nodes are expected to share similar labels.

Subsequent work extended PCC to several graph-based learning tasks, including mechanisms to reduce error propagation from mislabeled data in semi-supervised learning \citep{breve2011preventing} and fuzzy community detection \citep{breve2013fuzzy}. PCC was later explicitly adapted to handle label noise in semi-supervised learning through graph-based relabeling dynamics, reinforcing its robustness to corrupted labels \citep{breve2015particle}. More recent studies have also explored PCC in hybrid graph-learning settings. For example, \citet{leticio2025graph} use PCC-generated representations together with node attributes to improve graph convolutional networks, while \citet{pereira2026clustering} employ PCC as a graph-based label correction mechanism in structured data. Together, these studies suggest that PCC can serve as an effective graph-aware mechanism for identifying and correcting suspicious labels.

In the present work, PCC is used as a pre-training label refinement stage before GCN optimization. Given an initially labeled graph, PCC identifies suspicious labels through domination dynamics and performs confidence-aware label refinement, either relabeling nodes or removing uncertain labels when the inferred class support is insufficient. This process reduces the impact of noisy labels before GCN training. Unlike \citet{pereira2026clustering}, which applies PCC after clustering, our approach uses PCC directly as a denoising mechanism prior to GCN training.

\subsection{Instance-Dependent Noise in Graphs}

Most early work on noisy labels in graph learning focused on synthetic corruption processes such as uniform or class-dependent noise. Although useful for controlled evaluation, these settings do not always reflect the more complex forms of corruption found in real data. More recent research has therefore emphasized instance-dependent label noise, in which the probability of corruption depends on the characteristics of the individual sample itself \citep{yao2021instance,garg2023instance,kim2025delving}.

This setting is particularly relevant for graph learning because node attributes, local graph neighborhoods, and structural ambiguity may all influence the likelihood of mislabeling. In the present work, instance-dependent noise is used as an additional evaluation setting to assess whether PCC-based label refinement remains effective when label corruption is correlated with the input structure rather than generated through synthetic corruption processes.

\section{Proposed PCC+GCN Framework}\label{sec:ProposedFramework}

This section presents the proposed PCC+GCN framework for robust node classification under noisy labels. The framework consists of three stages. First, the graph used by PCC may be enhanced through the incorporation of additional k-nearest-neighbor (k-NN) edges derived from node features. Next, PCC performs graph-based label refinement by identifying suspicious labels and either relabeling or removing them according to a confidence criterion. Finally, a GCN is trained on the original graph structure and node features using the labels refined by PCC.

\subsection{Framework Overview}
\label{sec:FrameworkOverview}

\begin{figure}
	\centering
	\includegraphics[width=1\linewidth]{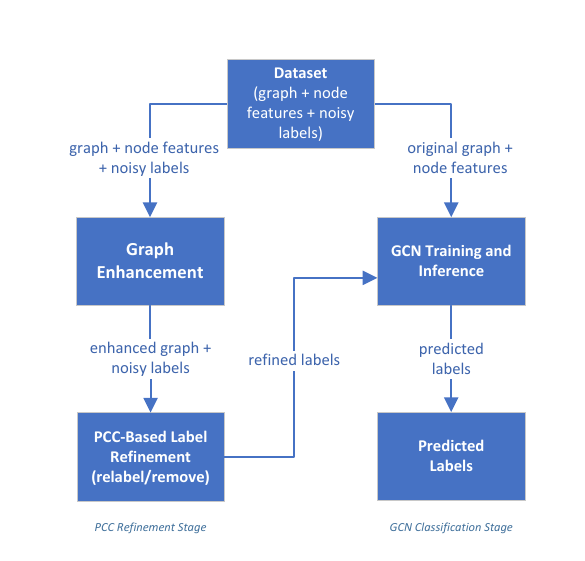}
	\caption[The proposed framework.]{The proposed framework.}
	\label{fig:pipeline}
\end{figure}

The framework is depicted in Figure~\ref{fig:pipeline}. As illustrated, the proposed framework separates label refinement from node classification. PCC operates on an enhanced graph to identify and correct potentially corrupted labels, while the GCN is trained on the original graph and node features using the refined labels generated by PCC. This design allows the denoising stage to be performed independently of the graph neural network architecture.

\subsection{Graph Enhancement}
\label{sec:GraphEnhancement}

Let $G=(V,E)$ denote the graph provided by the dataset, where $V$ is the set of nodes and $E$ is the set of edges. Each node $v \in V$ is associated with a feature vector $x_v$. Before applying PCC, the original graph may be augmented with additional edges obtained through a $k$-nearest-neighbor ($k$-NN) search in the feature space.

Four graph construction modes are considered for the PCC refinement stage. One mode uses the original graph without modification, whereas the remaining three modes augment the graph with additional $k$-NN edges according to different label-consistency criteria, as shown in Table~\ref{tab:GraphModes}.

\begin{table}[ht]
	\centering
	\caption{k-NN augmentation modes evaluated in the proposed framework.}
	\label{tab:GraphModes}
	\begin{tabular}{p{2cm} p{5cm}}
		\hline
		\textbf{Mode} & \textbf{Description} \\
		\hline
		None &
		Original graph without additional $k$-NN edges. \\
		
		Same-Label &
		Add a $k$-NN edge only when both endpoints are labeled and share the same class label. \\
		
		Non-Conflicting &
		Add a $k$-NN edge unless the endpoints have conflicting known labels. This includes pairs involving unlabeled nodes. \\
		
		Full &
		Add all $k$-NN edges regardless of label information. \\
		\hline
	\end{tabular}
\end{table}

The enhanced graph is used exclusively during the PCC refinement stage. The GCN is subsequently trained on the original graph provided by the dataset.

\subsection{PCC-Based Label Refinement}
\label{sec:PCCLabelRefinement}

The proposed framework employs the PCC model as a graph-based label refinement mechanism. PCC was originally introduced as a semi-supervised learning algorithm based on competition and cooperation dynamics among particles associated with different classes \citep{breve2012particle}. A detailed description of the model and its dynamics can be found in the original work and its subsequent extensions \citep{breve2012particle,breve2015particle}.

In the proposed framework, PCC is not used as the final classifier. Instead, it operates as a preprocessing stage that analyzes the enhanced graph and estimates the class support for each labeled node. Nodes whose original labels are inconsistent with the local domination dynamics are considered suspicious and become candidates for refinement.

In this work, the PCC variant proposed by \citet{breve2015particle} is adopted with two modifications. First, the original formulation based on separate random and greedy walk transition probabilities is employed, corresponding to Equations (2) and (3) in \citet{breve2015particle}, rather than the unified formulation presented in Equation (4). Consequently, the greedy-walk probability $p_{\mathrm{grd}}$ is treated as a tunable hyperparameter.

Second, the distance exponent used in the greedy transition rule is generalized. Whereas the original formulation employs a fixed quadratic distance term, the proposed framework introduces a distance exponent hyperparameter, denoted by $d_{\mathrm{exp}}$, allowing the influence of distance on particle movement to be tuned during hyperparameter optimization.

Let $v_i^{\lambda_c}$ denote the accumulated domination level of class $c$ over node $i$, as defined in \citet{breve2015particle}. In the proposed framework, these accumulated domination levels are interpreted as class-support scores and are used to determine whether a labeled node should be preserved, removed, or relabeled. Although accumulated domination levels are not probabilities, they provide a natural measure of the support that the PCC dynamics assigns to each class.

Two confidence thresholds are introduced. The removal threshold, denoted by $\tau_{\mathrm{rem}}$, determines whether the original label remains sufficiently supported by the PCC dynamics. If the accumulated domination level associated with the node's original class falls below $\tau_{\mathrm{rem}}$, the label is removed and the node becomes unlabeled.

The relabeling threshold, denoted by $\tau_{\mathrm{rel}}$, controls when a node may receive a new label. If the accumulated domination level of the most dominant class (i.e., the class with the highest accumulated domination level) exceeds $\tau_{\mathrm{rel}}$ and differs from the original class, the node is relabeled accordingly. Otherwise, the original label is preserved whenever it satisfies the removal criterion.

\subsection{GCN Training and Inference}
\label{sec:GCNTraining}

After the PCC refinement stage, a Graph Convolutional Network (GCN) \citep{kipf2017semi} is trained using the original graph structure provided by the dataset together with the original node features. The graph enhancement is used exclusively during the PCC refinement stage; the GCN is trained on the original graph, which naturally combines graph topology and node features during message passing.

The refined labels produced by PCC are used as supervision during training. Nodes whose labels are preserved or relabeled contribute to the supervised loss according to their final assigned class, whereas nodes whose labels are removed are treated as unlabeled and therefore do not contribute to the loss function.

During inference, the trained GCN predicts the labels of all nodes following the standard semi-supervised node classification procedure proposed by \citet{kipf2017semi}. No additional post-processing or label refinement is performed after GCN inference.

\section{Experimental Setup} \label{sec:ExperimentalSetup}

This section describes the experimental setup adopted throughout this work. It first presents the datasets and label-noise models used in the experiments. Next, the evaluation protocol is described, including the adopted performance metric and comparison methodology. Finally, the implementation details, computational environment, and hyperparameter optimization procedures are provided.

\subsection{Datasets}

The experiments were conducted on the benchmark datasets provided by the NoisyGL framework \citep{wang2024noisygl}. This benchmark comprises graph datasets with different sizes, feature dimensions, class distributions, and levels of homophily, making it suitable for evaluating graph neural networks under label noise.

The hyperparameter analysis was conducted on the Cora, CiteSeer, and PubMed datasets. These datasets were selected to analyze the sensitivity of the proposed framework to different hyperparameter settings through exhaustive grid-search heatmaps, thereby providing insight into how each hyperparameter influences classification performance.

The benchmark comparison and the instance-dependent label-noise experiments were conducted on the complete set of datasets available in the NoisyGL benchmark.

Table~\ref{tab:Datasets} summarizes the characteristics of the datasets used throughout this work.

\begin{table*}[t]
	\centering
	\caption{Characteristics of the datasets used in this work (adapted from \citet{wang2024noisygl}).}
	\label{tab:Datasets}
	\begin{tabular}{lrrrrrrc}
		\toprule
		\textbf{Dataset} &
		\textbf{\# Nodes} &
		\textbf{\# Edges} &
		\textbf{Features} &
		\textbf{\# Classes} &
		\textbf{Homophily} &
		\textbf{Avg. Degree} &
		\textbf{HP Analysis} \\
		\midrule
		Cora               & 2,708  & 5,278   & 1,433  & 7  & 0.81 & 3.90 & \checkmark \\
		CiteSeer           & 3,327  & 4,552   & 3,703  & 6  & 0.74 & 2.74 & \checkmark \\
		PubMed             & 19,717 & 44,324  & 500    & 3  & 0.80 & 4.50 & \checkmark \\
		Amazon-Computers   & 13,752 & 491,722 & 767    & 10 & 0.78 & 35.8 & \\
		Amazon-Photos      & 7,650  & 238,162 & 745    & 8  & 0.83 & 31.1 & \\
		DBLP               & 17,716 & 105,734 & 1,639  & 4  & 0.83 & 5.97 & \\
		BlogCatalog        & 5,196  & 343,486 & 8,189  & 6  & 0.40 & 66.1 & \\
		Flickr             & 7,575  & 239,738 & 12,047 & 9  & 0.24 & 63.3 & \\
		Amazon-Ratings     & 24,492 & 93,050  & 300    & 5  & 0.38 & 7.60 & \\
		Roman-Empire       & 22,662 & 32,927  & 300    & 18 & 0.05 & 2.90 & \\
		\bottomrule
	\end{tabular}
\end{table*}

\subsection{Noise Models}

Three experimental settings were considered in this work: (i) hyperparameter analysis, (ii) benchmark evaluation under conventional label-noise models, and (iii) evaluation under instance-dependent label noise.

For the hyperparameter analysis, Uniform label noise was adopted at corruption rates ranging from $10\%$ to $50\%$ in increments of $10\%$ over the initially labeled nodes. This setting allowed the sensitivity of the PCC hyperparameters to be evaluated across different levels of label corruption while keeping the noise model fixed.

For the benchmark comparison, the same conventional label-noise models adopted by the NoisyGL benchmark \citep{wang2024noisygl} were considered, namely Uniform, Pair, and Random label noise. Following the benchmark protocol, corruption rates ranging from 10\% to 50\% were evaluated. Only the proposed PCC+GCN framework was executed for the conventional label-noise benchmark. The results of the competing methods were taken directly from the NoisyGL benchmark.

Finally, the proposed framework was evaluated under instance-dependent label noise, in which the probability of label corruption depends on the characteristics of each individual node rather than being generated uniformly at random. The instance-dependent noise model follows the implementation provided by the NoisyGL framework. In this work, only its implementation was redesigned to exploit GPU acceleration, while preserving the original noise-generation procedure. This setting provides a more realistic and challenging evaluation scenario because the label noise is correlated with the underlying data distribution \citep{yao2021instance,garg2023instance,kim2025delving}.

\subsection{Evaluation Protocol}

Classification performance was evaluated using node classification accuracy as the primary metric. All reported results correspond to the mean accuracy and standard deviation over 10 independent runs.

For the hyperparameter analysis, the average accuracy obtained over the 10 runs was used to generate the heatmaps, allowing the influence of each hyperparameter on the classification performance of the proposed framework to be analyzed.

For the conventional label-noise benchmark, the original train/validation/test splits and evaluation protocol provided by NoisyGL \citep{wang2024noisygl} were adopted to ensure a fair comparison with previously published methods. In addition to the individual dataset results, the overall performance was assessed through the average ranking of the methods across all datasets, where lower average ranks indicate better overall performance.

For the instance-dependent label-noise experiments, the same evaluation protocol was adopted. Since all methods were executed within the same experimental framework, computational cost was also evaluated by measuring the execution time of both CPU and GPU stages independently.

\subsection{Implementation Details}
\label{sec:ImplementationDetails}

The proposed framework was implemented in Python using the PyTorch and PyTorch Geometric libraries. The benchmark and instance-dependent label-noise experiments were implemented by extending the NoisyGL framework \citep{wang2024noisygl}.

For the hyperparameter analysis, the PCC hyperparameters were systematically evaluated through an exhaustive grid search to analyze the sensitivity of the proposed framework. For these experiments, the GCN configuration was kept fixed, using a two-layer GCN with 16 hidden units, a dropout rate of $0.5$, a learning rate of $0.01$, a weight decay of $5\times10^{-4}$, and 200 training epochs. During GCN training, the model state achieving the highest validation accuracy was retained and subsequently used to evaluate the test set.

The accumulated-domination increment was fixed at $\Delta_v=0.1$ throughout all experiments. The internal restart parameter of PCC was fixed at 10. At each restart, the particle competition process is reinitialized while the accumulated domination levels are retained. This strategy, originally proposed by \citet{breve2015particle}, reduces the influence of stochastic particle movements on the final accumulated domination levels.

Only the proposed PCC+GCN method was executed for the conventional benchmark. The hyperparameters of the proposed method were optimized using Optuna \citep{akiba2019optuna}, whereas the competing methods were evaluated using the results reported by the NoisyGL benchmark \citep{wang2024noisygl}, which were obtained using the benchmark's original optimization procedure.

For the instance-dependent label-noise experiments, all baseline methods and the proposed method were executed within the extended NoisyGL framework. For all methods, the hyperparameters were optimized using Optuna. The optimized GCN hyperparameters included the number of hidden units, the number of graph convolutional layers, the learning rate, the weight decay, and the dropout rate.

The original NoisyGL implementation already includes an instance-dependent label-noise generator, although it was not used in the benchmark experiments reported in \citet{wang2024noisygl}. Since all compared methods had to be executed under this noise model in the present work, the generator was redesigned to improve its computational efficiency while preserving the original noise-generation procedure. The optimized implementation groups nodes according to their class labels and performs projection and sampling operations in batches using GPU-accelerated tensor operations. Instead of processing nodes individually, matrix multiplications and multinomial sampling are executed in parallel through PyTorch, while class-wise processing avoids the excessive memory consumption of a fully vectorized implementation. These modifications substantially reduced preprocessing time, making large-scale instance-dependent label-noise experiments computationally feasible.

Unlike the conventional benchmark, where only the proposed PCC+GCN method was executed, all methods compared under instance-dependent label noise were evaluated using the same implementation and hyperparameter optimization procedure, ensuring a fair comparison.

All experiments were performed on a workstation equipped with an Intel Core i9-14900K processor, 128 GB of RAM, an NVIDIA GeForce RTX 4060 Ti GPU, and an NVIDIA GeForce RTX 2060 SUPER GPU.

\section{Hyperparameter Analysis} \label{sec:HyperparameterAnalysis}

This section presents the analysis of the PCC hyperparameters on three datasets: Cora, CiteSeer, and PubMed, under noise rates ranging from $10\%$ to $50\%$ in increments of $10\%$. The GCN configuration was kept fixed throughout these experiments, as described in Section~\ref{sec:ImplementationDetails}. Throughout this section, accuracy gain denotes the difference in test accuracy between the proposed PCC+GCN pipeline and the GCN baseline trained directly on the noisy labels.

\subsection{Greedy-Walk Probability and Distance Exponent}

In the first experiment, the greedy-walk probability $p_{\mathrm{grd}}$ was varied over
$p_{\mathrm{grd}} \in \{0.0, 0.1, 0.2, \ldots, 0.9\}$,
while the distance exponent was varied over
$d_{\mathrm{exp}} \in \{0, 1, 2, \ldots, 10\}$.
The removal and relabeling thresholds were fixed at
$\tau_{\mathrm{rem}} = 0.5$ and
$\tau_{\mathrm{rel}} = 0.5$, respectively, and no $k$-NN graph augmentation was applied (i.e., the ``None'' graph construction mode).

\begin{figure*}
	\centering
	\includegraphics[width=1\linewidth]{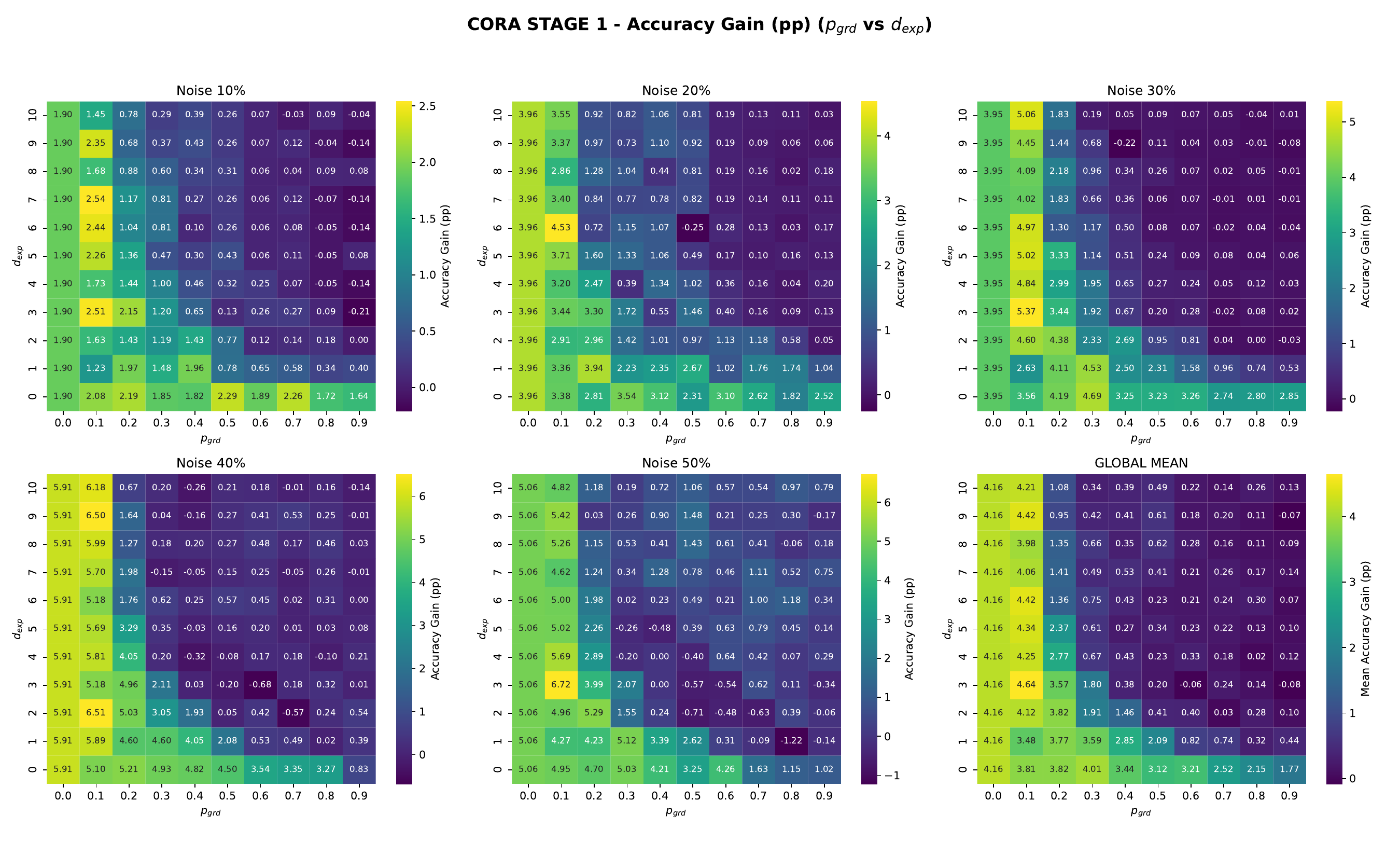}
	\caption{Heatmaps for the Cora dataset with different noise rates, greedy-walk probabilities, and distance exponents.}
	\label{fig:heatmapsstage1cora}
\end{figure*}

\begin{figure*}
	\centering
	\includegraphics[width=1\linewidth]{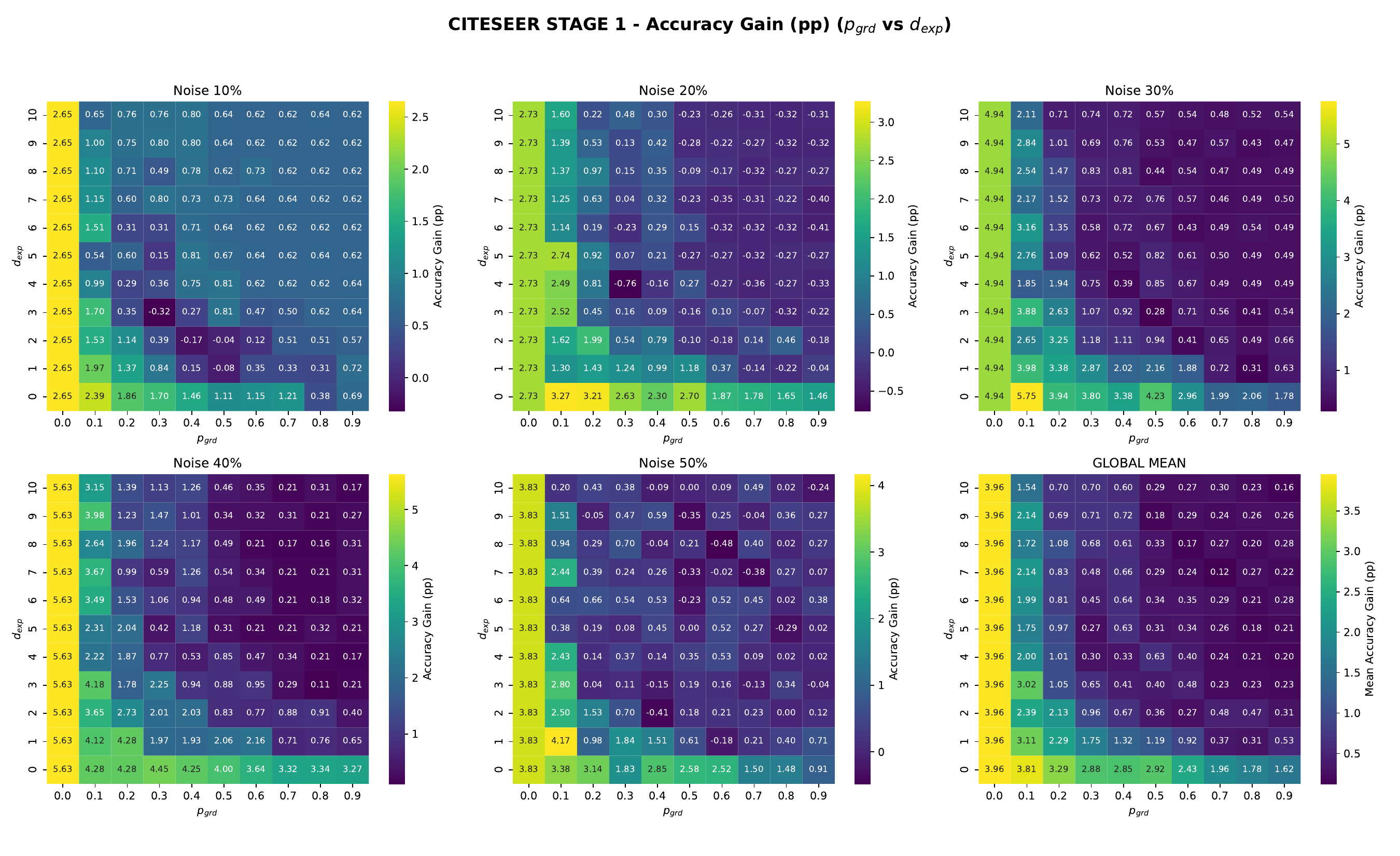}
	\caption{Heatmaps for the CiteSeer dataset with different noise rates, greedy-walk probabilities, and distance exponents.}
	\label{fig:heatmapsstage1citeseer}
\end{figure*}

\begin{figure*}
	\centering
	\includegraphics[width=1\linewidth]{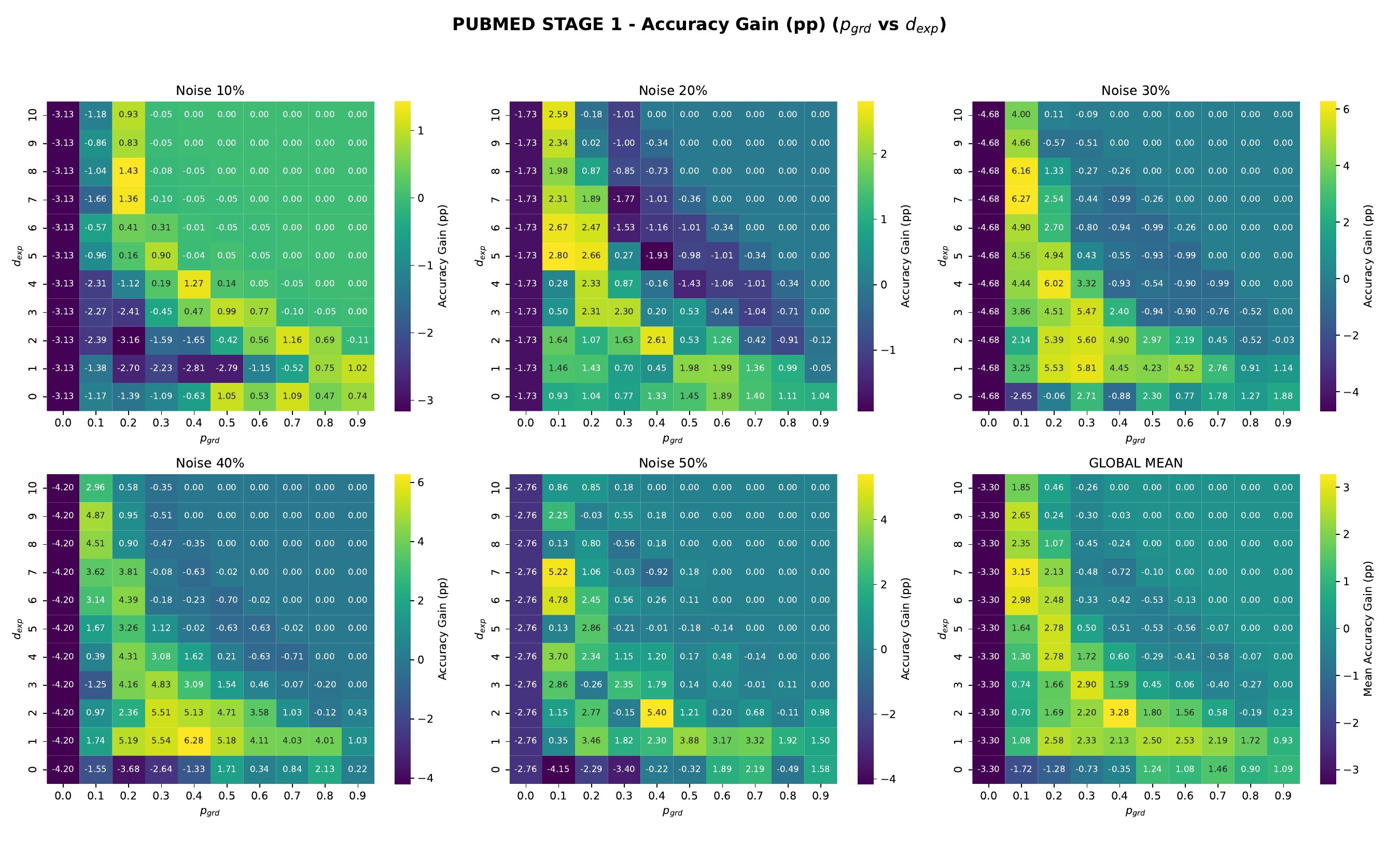}
	\caption{Heatmaps for the PubMed dataset with different noise rates, greedy-walk probabilities, and distance exponents.}
	\label{fig:heatmapsstage1pubmed}
\end{figure*}

Figures~\ref{fig:heatmapsstage1cora}, \ref{fig:heatmapsstage1citeseer}, and \ref{fig:heatmapsstage1pubmed} show the heatmaps obtained for Cora, CiteSeer, and PubMed, respectively. The heatmaps report the accuracy gain for each combination of $p_{\mathrm{grd}}$ and $d_{\mathrm{exp}}$ at each noise rate.

For the Cora dataset, the highest accuracy gain is $6.72$ percentage points, observed at a noise rate of $50\%$. The value $p_{\mathrm{grd}}=0.1$ provides the best performance across all noise rates, while larger values consistently lead to lower accuracy gains. In contrast, $d_{\mathrm{exp}}$ shows considerably lower sensitivity, with relatively small variations in performance across its range. The highest average performance is obtained with $d_{\mathrm{exp}}=3$.

For the CiteSeer dataset, the highest accuracy gain is $5.75$ percentage points, observed at a noise rate of $30\%$. The value $p_{\mathrm{grd}}=0.0$ provides the best performance across all noise rates, with larger values consistently degrading the results. This indicates that, for this dataset, relying exclusively on random walks without greedy walks is more effective. When $p_{\mathrm{grd}}=0.0$, the distance exponent has no effect on the model. For $p_{\mathrm{grd}}>0.0$, the best performance is obtained with $d_{\mathrm{exp}}=0$, suggesting that, for this dataset, it is preferable not to emphasize previously established domination levels through the dominance- and distance-dependent greedy walk.

Finally, for the PubMed dataset, the highest accuracy gain is $6.28$ percentage points, observed at a noise rate of $40\%$. The results exhibit a clear trade-off between $p_{\mathrm{grd}}$ and $d_{\mathrm{exp}}$: higher values of $p_{\mathrm{grd}}$ tend to be associated with lower optimal values of $d_{\mathrm{exp}}$, and vice versa. The highest average performance is obtained with $p_{\mathrm{grd}}=0.4$ and $d_{\mathrm{exp}}=2$.

\subsection{Label Refinement Thresholds}

In the second experiment, the removal threshold was varied over
$\tau_{\mathrm{rem}} \in \{0.1, 0.2, 0.3, \ldots, 1.0\}$,
while the relabeling threshold was varied over
$\tau_{\mathrm{rel}} \in \{0.0, 0.1, 0.2, \ldots, 1.0\}$.
The greedy-walk probability and distance exponent were fixed for each dataset at the best values obtained in the first experiment, as shown in Table~\ref{tab:best_pcc_hyperparameters}. As in the first experiment, no $k$-NN graph augmentation was applied (i.e., the ``None'' graph construction mode).

\begin{table}[t]
	\centering
	\caption{PCC hyperparameter configurations selected from the hyperparameter analysis.}
	\label{tab:best_pcc_hyperparameters}
	\begin{tabular}{lrrrr}
		\toprule
		\textbf{Dataset} &
		$\boldsymbol{p_{\mathrm{grd}}}$ &
		$\boldsymbol{d_{\mathrm{exp}}}$ &
		$\boldsymbol{\tau_{\mathrm{rem}}}$ &
		$\boldsymbol{\tau_{\mathrm{rel}}}$ \\
		\midrule
		Cora     & 0.1 & 3 & 0.1 & 0.1 \\
		CiteSeer & 0.0 & 0 & 0.1 & 0.1 \\
		PubMed   & 0.4 & 2 & 0.1 & 0.1 \\
		\bottomrule
	\end{tabular}
\end{table}

\begin{figure*}
	\centering
	\includegraphics[width=1\linewidth]{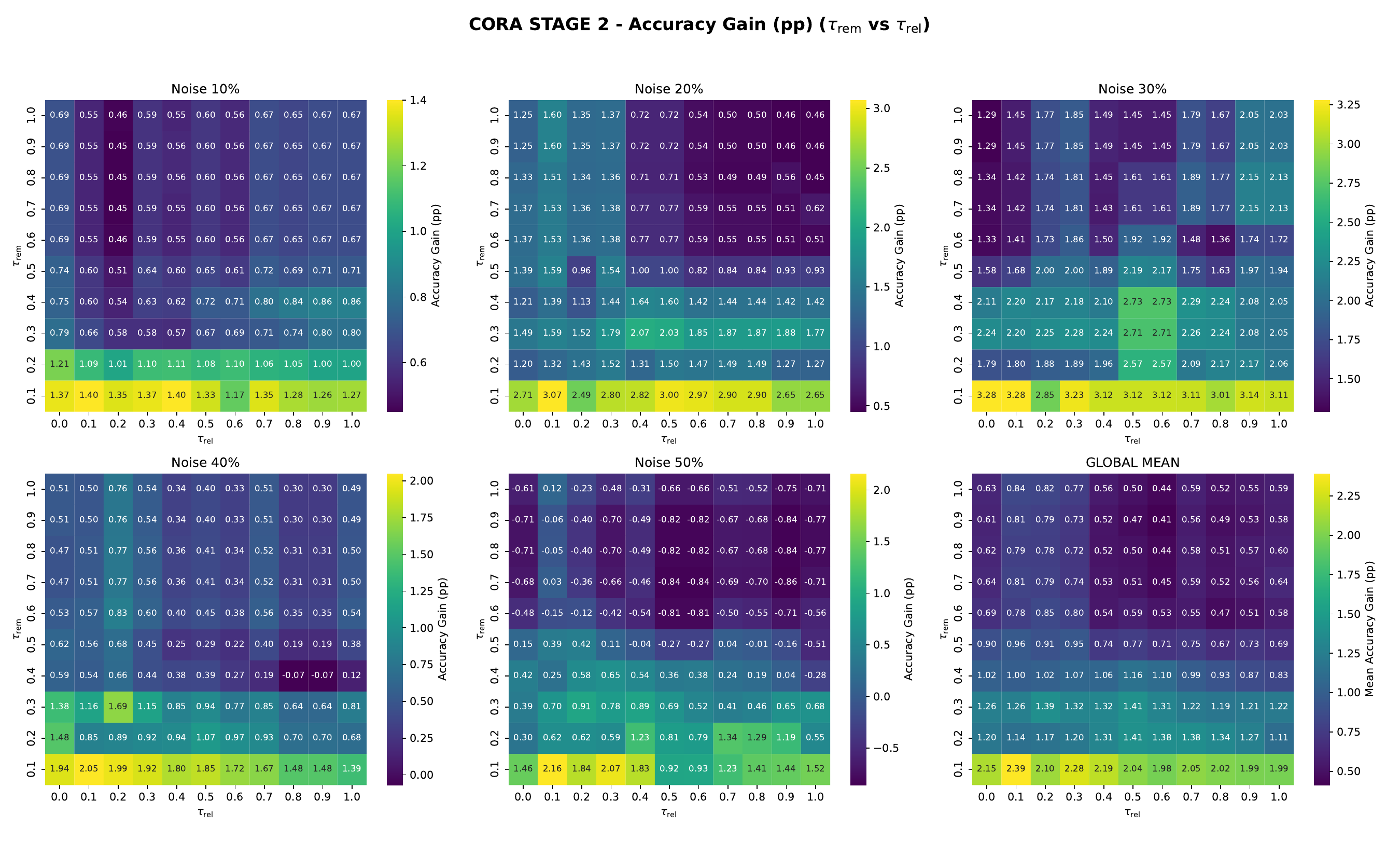}
	\caption{Heatmaps for the Cora dataset with different noise rates, removal and relabeling thresholds.}
	\label{fig:heatmapsstage2cora}
\end{figure*}

\begin{figure*}
	\centering
	\includegraphics[width=1\linewidth]{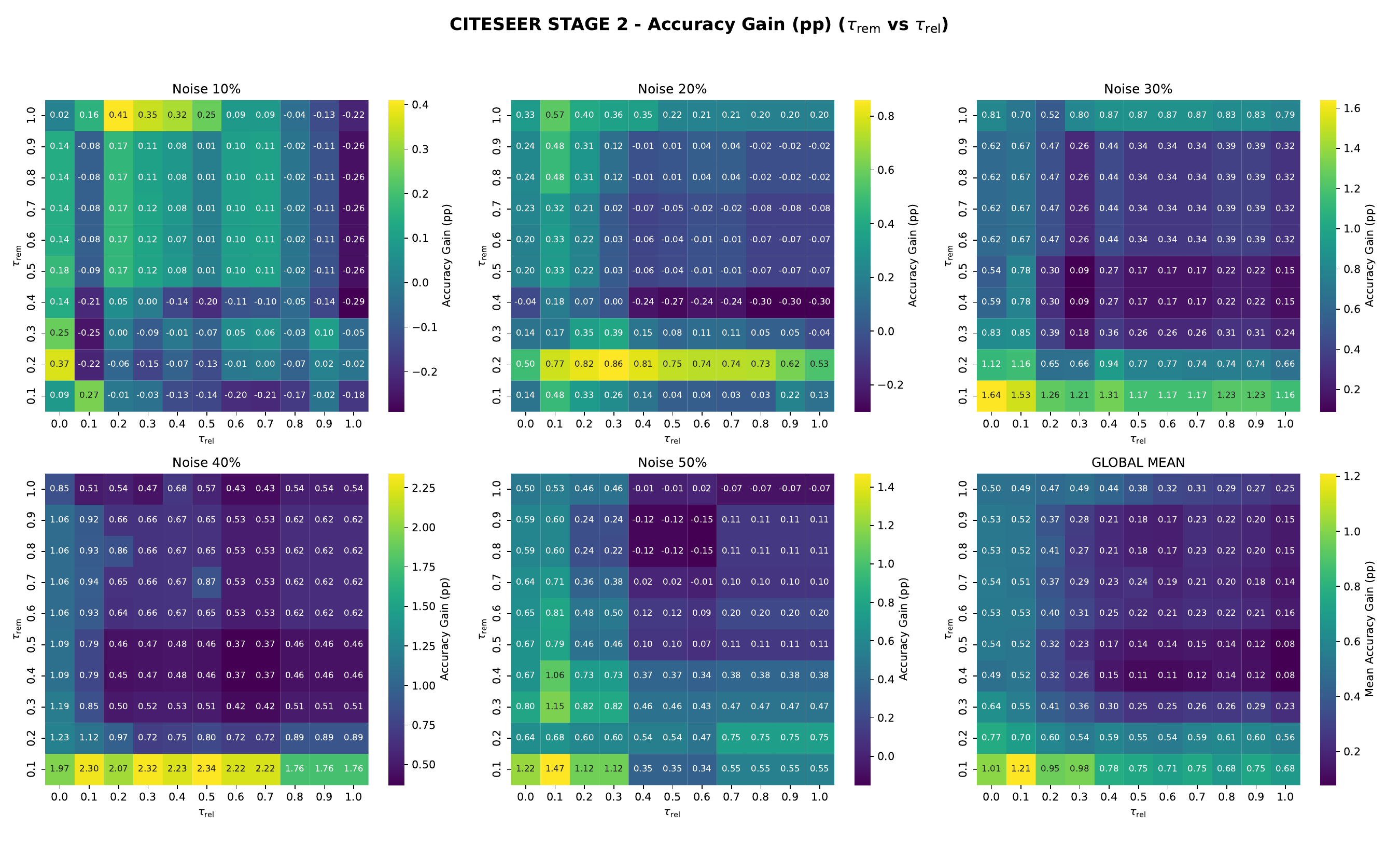}
	\caption{Heatmaps for the CiteSeer dataset with different noise rates, removal and relabeling thresholds.}
	\label{fig:heatmapsstage2citeseer}
\end{figure*}

\begin{figure*}
	\centering
	\includegraphics[width=1\linewidth]{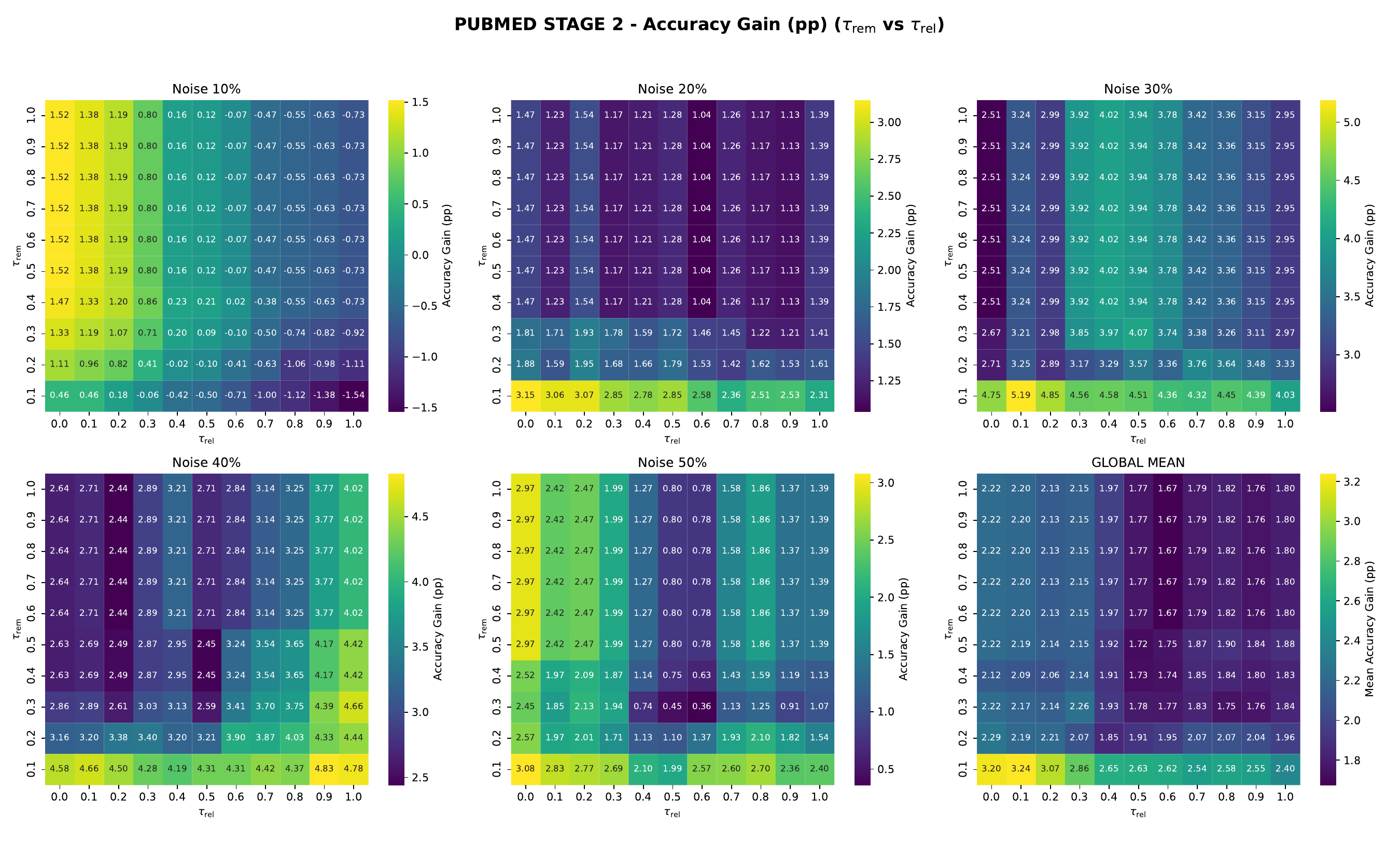}
	\caption{Heatmaps for the PubMed dataset with different noise rates, removal and relabeling thresholds.}
	\label{fig:heatmapsstage2pubmed}
\end{figure*}

Figures~\ref{fig:heatmapsstage2cora}, \ref{fig:heatmapsstage2citeseer}, and \ref{fig:heatmapsstage2pubmed} show the heatmaps obtained for Cora, CiteSeer, and PubMed, respectively. The heatmaps report the accuracy gain for each combination of $\tau_{\mathrm{rem}}$ and $\tau_{\mathrm{rel}}$ at each noise rate.

For the Cora dataset, the highest accuracy gain is $3.28$ percentage points, observed at a noise rate of $30\%$. The values $\tau_{\mathrm{rem}}=\tau_{\mathrm{rel}}=0.1$ provide the best performance across all noise rates. In general, the removal threshold is more sensitive than the relabeling threshold for this dataset.

For the CiteSeer dataset, the highest accuracy gain is $2.34$ percentage points, observed at a noise rate of $40\%$. Again, the values $\tau_{\mathrm{rem}}=\tau_{\mathrm{rel}}=0.1$ provide the best average performance, but for this dataset the optimal threshold pair varies across noise rates. For example, at a noise rate of $10\%$, $\tau_{\mathrm{rem}}=1.0$ and $\tau_{\mathrm{rel}}=0.2$ achieve the highest accuracy gain. At a noise rate of $20\%$, the best combination is $\tau_{\mathrm{rem}}=0.2$ and $\tau_{\mathrm{rel}}=0.3$.

Finally, for the PubMed dataset, the highest accuracy gain is $5.19$ percentage points, observed at a noise rate of $30\%$. Once again, $\tau_{\mathrm{rem}}=\tau_{\mathrm{rel}}=0.1$ provides the best average performance, while the optimal threshold pair varies across noise rates. For a noise rate of $10\%$, $\tau_{\mathrm{rel}}=0.0$ with $\tau_{\mathrm{rem}} \geq 0.5$ achieves the best results. For noise rates of $20\%$ or higher, $\tau_{\mathrm{rem}}=0.1$ with $\tau_{\mathrm{rel}} \leq 0.1$ achieves the best results.

\subsection{$k$-NN Graph Augmentation}

In the third experiment, the three $k$-NN graph augmentation modes were evaluated together with the original graph. The number of nearest neighbors was varied over
$k \in \{2, 5, 10, 15, 20, 30, 50, 75, 100\}$,
while the other PCC hyperparameters were fixed for each dataset at the best values obtained in the previous experiments, as summarized in Table~\ref{tab:best_pcc_hyperparameters}.

\begin{figure*}
	\centering
	\includegraphics[width=1\linewidth]{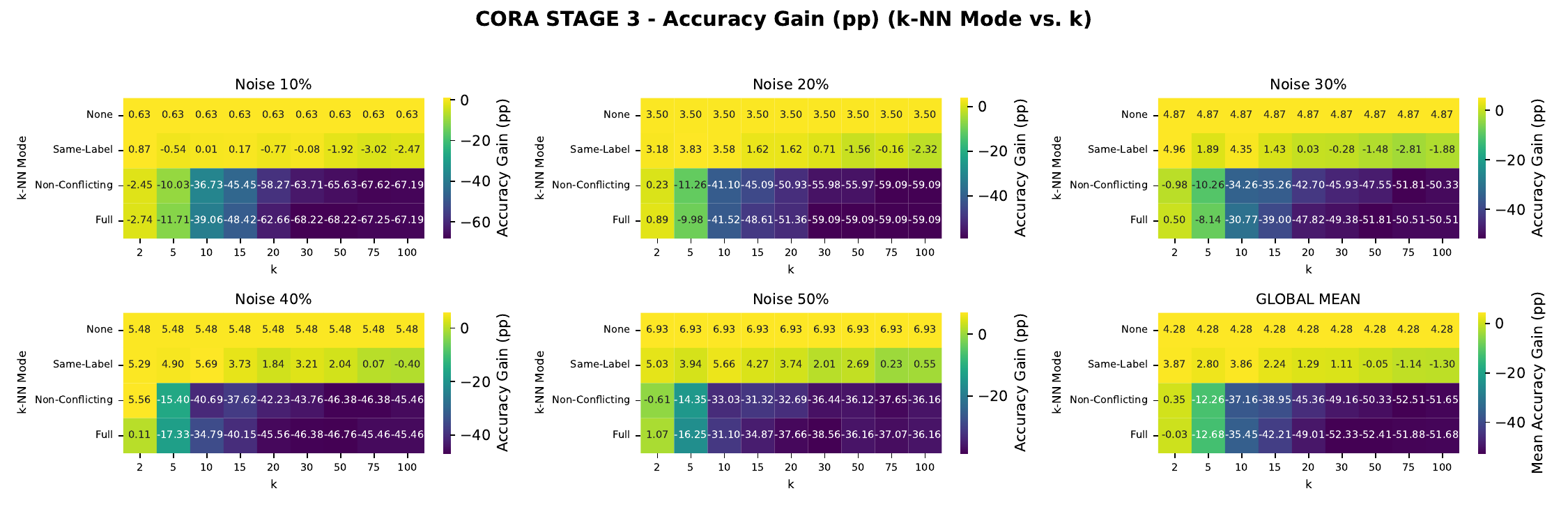}
	\caption{Heatmaps for the Cora dataset with different noise rates, graph augmentation modes, and numbers of nearest-neighbor edges added.}
	\label{fig:heatmapsstage3cora}
\end{figure*}

\begin{figure*}
	\centering
	\includegraphics[width=1\linewidth]{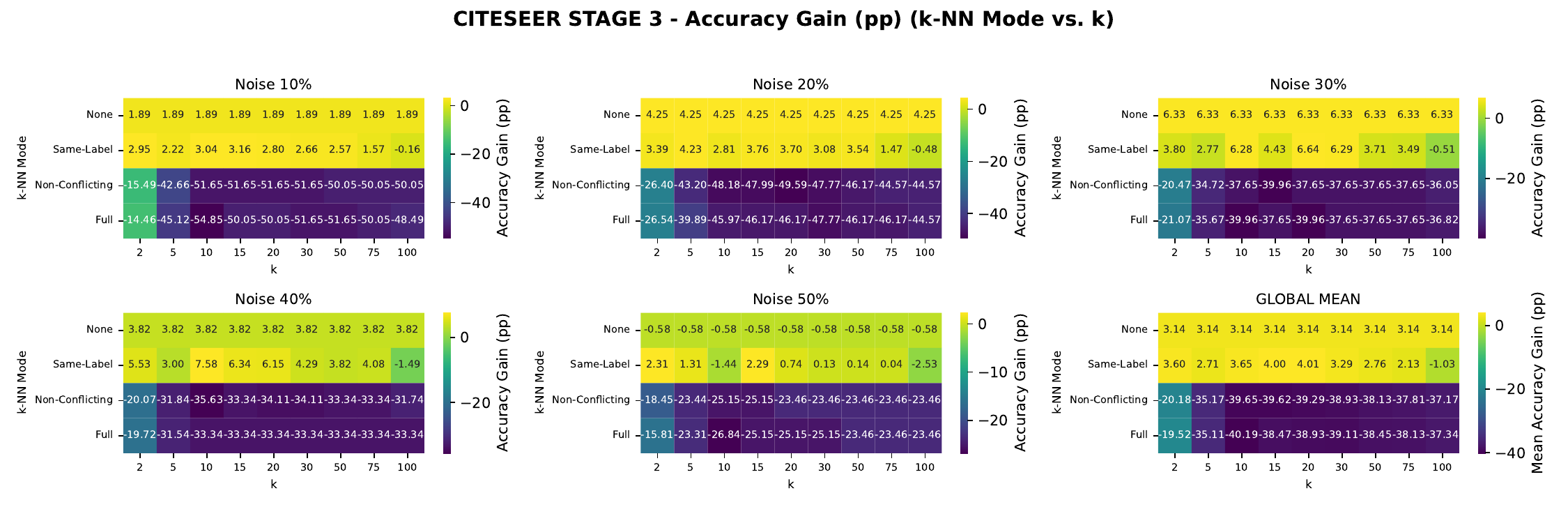}
	\caption{Heatmaps for the CiteSeer dataset with different noise rates, graph augmentation modes, and numbers of nearest-neighbor edges added.}
	\label{fig:heatmapsstage3citeseer}
\end{figure*}

\begin{figure*}
	\centering
	\includegraphics[width=1\linewidth]{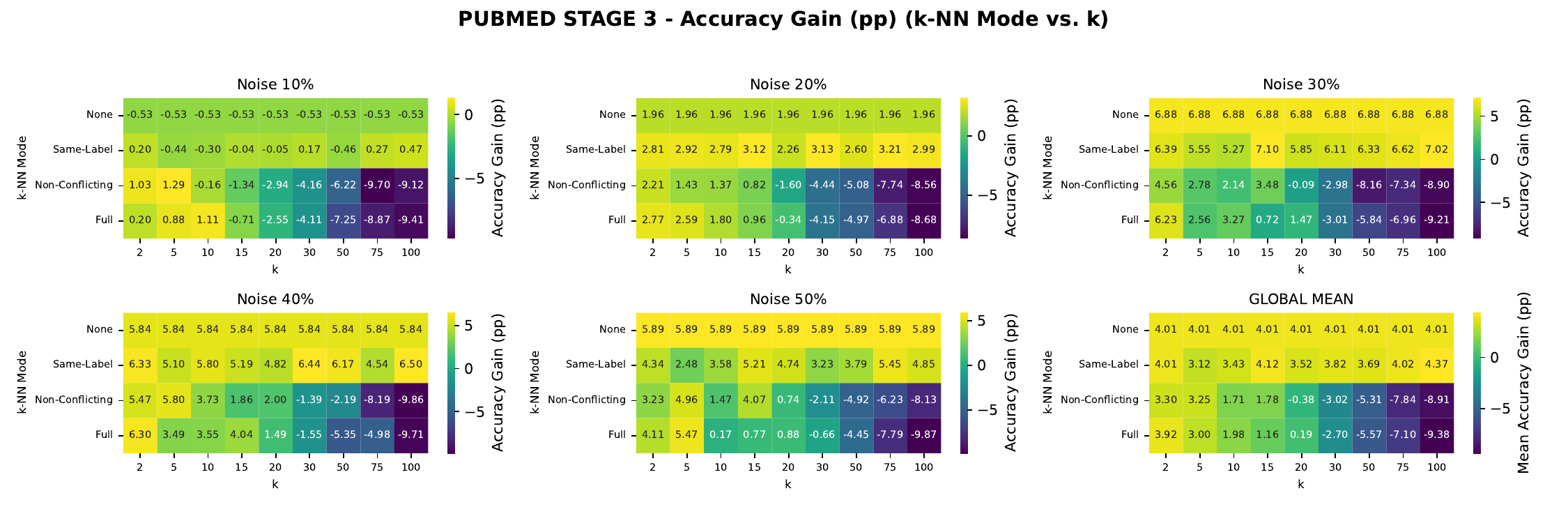}
	\caption{Heatmaps for the PubMed dataset with different noise rates, graph augmentation modes, and numbers of nearest-neighbor edges added.}
	\label{fig:heatmapsstage3pubmed}
\end{figure*}

Figures~\ref{fig:heatmapsstage3cora}, \ref{fig:heatmapsstage3citeseer}, and \ref{fig:heatmapsstage3pubmed} show the heatmaps obtained for Cora, CiteSeer, and PubMed, respectively. The heatmaps report the accuracy gain for each graph construction mode and value of $k$ at each noise rate.

For the Cora dataset, the highest accuracy gain is $6.93$ percentage points, observed at a noise rate of $50\%$ with no graph augmentation. Among the augmentation strategies, only the ``Same-Label'' strategy improved the accuracy gain across all scenarios with noise rates of $40\%$ or lower. In all these cases, the best values of $k$ were small, ranging from $k=2$ to $k=10$.

For the CiteSeer dataset, the highest accuracy gain is $7.58$ percentage points, observed at a noise rate of $40\%$ with no graph augmentation. Again, only the ``Same-Label'' strategy improved the accuracy gain. On average, $k=20$ led to the best results, with the best-performing range lying between $k=10$ and $k=30$.

Finally, for the PubMed dataset, the highest accuracy gain is $7.10$ percentage points, observed at a noise rate of $30\%$. Unlike the other datasets, all graph augmentation strategies improved the accuracy gain in some scenarios. However, the best results were still achieved with the ``Same-Label'' strategy, with $k=100$.

It is worth noting that the ``Same-Label'' strategy achieved its best results across practically the entire range of $k$ values tested, whereas the other two strategies achieved their best average results at $k=2$. This behavior is expected because relatively few candidate edges satisfy the ``Same-Label'' criterion, while the ``Non-Conflicting'' strategy accepts most candidate edges and the ``Full'' strategy accepts all candidate edges.

\section{Benchmark Comparison} \label{sec:BenchmarkComparison}

This section compares the proposed PCC+GCN method with the methods included in the NoisyGL benchmark \citep{wang2024noisygl} across the ten datasets described in Table~\ref{tab:Datasets}. The comparison considers the three label-noise models adopted by the benchmark, namely Uniform, Pair, and Random, at noise rates ranging from $10\%$ to $50\%$. In addition, the clean setting was included as a reference without label corruption.

Table~\ref{tab:BenchmarkAbsResults} reports the test accuracy of the proposed method for all datasets, noise types, and noise rates. Table~\ref{tab:BenchmarkGainResults} reports the corresponding accuracy gain relative to the baseline GCN, expressed in percentage points. The largest individual gain was observed on the CiteSeer dataset under $40\%$ Uniform noise, where the proposed method achieved an accuracy gain of $11.67$ percentage points. On average, the largest gain was observed on CiteSeer, with $5.94$ percentage points. The proposed method achieved a positive average gain on nine of the ten datasets, with Amazon-Computers being the only exception, where the average accuracy decreased by $3.61$ percentage points. Overall, the average gain across the clean setting and all noisy scenarios was $1.67$ percentage points.

\begin{table*}
	\centering
	\small
	\caption{Accuracy (\%) of the proposed method in the NoisyGL Benchmark for all datasets, noise types and noise rates.}
	\label{tab:BenchmarkAbsResults}
	\resizebox{\textwidth}{!}{%
		\begin{tabular}{llcccccccccc}
			\toprule
			\textbf{Noise} & \textbf{Rate} & \textbf{Cora} & \textbf{CiteSeer} & \textbf{PubMed} & \textbf{Amazon-C} & \textbf{Amazon-P} & \textbf{DBLP} & \textbf{BlogCatalog} & \textbf{Flickr} & \textbf{Amz-Rat.} & \textbf{Roman-Emp.} \\
			\midrule
			Clean & Clean & $82.03{\scriptscriptstyle\pm}0.51$ & $72.53{\scriptscriptstyle\pm}0.89$ & $78.99{\scriptscriptstyle\pm}0.60$ & $82.16{\scriptscriptstyle\pm}0.54$ & $91.26{\scriptscriptstyle\pm}0.43$ & $78.71{\scriptscriptstyle\pm}0.41$ & $74.87{\scriptscriptstyle\pm}0.83$ & $59.75{\scriptscriptstyle\pm}0.79$ & $40.19{\scriptscriptstyle\pm}0.18$ & $42.63{\scriptscriptstyle\pm}0.24$ \\
			Uniform & 0.1 & $80.83{\scriptscriptstyle\pm}1.27$ & $70.39{\scriptscriptstyle\pm}1.37$ & $74.18{\scriptscriptstyle\pm}3.23$ & $81.38{\scriptscriptstyle\pm}0.94$ & $89.02{\scriptscriptstyle\pm}1.67$ & $76.25{\scriptscriptstyle\pm}1.32$ & $73.23{\scriptscriptstyle\pm}1.31$ & $56.60{\scriptscriptstyle\pm}0.94$ & $39.60{\scriptscriptstyle\pm}0.53$ & $40.75{\scriptscriptstyle\pm}0.74$ \\
			& 0.2 & $78.54{\scriptscriptstyle\pm}1.62$ & $67.93{\scriptscriptstyle\pm}2.32$ & $71.02{\scriptscriptstyle\pm}3.70$ & $77.13{\scriptscriptstyle\pm}2.23$ & $87.00{\scriptscriptstyle\pm}4.02$ & $73.75{\scriptscriptstyle\pm}2.87$ & $71.53{\scriptscriptstyle\pm}1.19$ & $54.64{\scriptscriptstyle\pm}1.66$ & $39.58{\scriptscriptstyle\pm}0.39$ & $38.53{\scriptscriptstyle\pm}0.63$ \\
			& 0.3 & $72.08{\scriptscriptstyle\pm}3.31$ & $63.93{\scriptscriptstyle\pm}3.19$ & $63.12{\scriptscriptstyle\pm}7.20$ & $74.20{\scriptscriptstyle\pm}3.17$ & $85.97{\scriptscriptstyle\pm}2.96$ & $68.82{\scriptscriptstyle\pm}2.70$ & $70.36{\scriptscriptstyle\pm}1.85$ & $51.71{\scriptscriptstyle\pm}2.00$ & $39.37{\scriptscriptstyle\pm}0.47$ & $36.66{\scriptscriptstyle\pm}0.48$ \\
			& 0.4 & $63.37{\scriptscriptstyle\pm}6.05$ & $60.56{\scriptscriptstyle\pm}4.10$ & $56.56{\scriptscriptstyle\pm}5.26$ & $68.98{\scriptscriptstyle\pm}9.04$ & $83.16{\scriptscriptstyle\pm}3.75$ & $66.94{\scriptscriptstyle\pm}4.72$ & $68.11{\scriptscriptstyle\pm}2.94$ & $47.86{\scriptscriptstyle\pm}1.76$ & $37.73{\scriptscriptstyle\pm}0.96$ & $33.17{\scriptscriptstyle\pm}0.75$ \\
			& 0.5 & $57.96{\scriptscriptstyle\pm}9.38$ & $51.98{\scriptscriptstyle\pm}4.84$ & $48.00{\scriptscriptstyle\pm}10.53$ & $62.47{\scriptscriptstyle\pm}7.78$ & $76.89{\scriptscriptstyle\pm}5.15$ & $62.75{\scriptscriptstyle\pm}5.77$ & $61.16{\scriptscriptstyle\pm}4.86$ & $43.51{\scriptscriptstyle\pm}2.70$ & $36.65{\scriptscriptstyle\pm}1.30$ & $33.63{\scriptscriptstyle\pm}0.66$ \\
			Pair & 0.1 & $79.59{\scriptscriptstyle\pm}1.84$ & $69.65{\scriptscriptstyle\pm}2.11$ & $75.87{\scriptscriptstyle\pm}1.94$ & $80.56{\scriptscriptstyle\pm}1.79$ & $87.73{\scriptscriptstyle\pm}2.56$ & $76.78{\scriptscriptstyle\pm}1.37$ & $72.75{\scriptscriptstyle\pm}1.61$ & $56.95{\scriptscriptstyle\pm}0.91$ & $40.12{\scriptscriptstyle\pm}0.27$ & $40.40{\scriptscriptstyle\pm}0.63$ \\
			& 0.2 & $67.04{\scriptscriptstyle\pm}9.36$ & $66.40{\scriptscriptstyle\pm}3.16$ & $71.56{\scriptscriptstyle\pm}2.07$ & $76.92{\scriptscriptstyle\pm}2.54$ & $84.07{\scriptscriptstyle\pm}3.42$ & $73.05{\scriptscriptstyle\pm}1.89$ & $70.84{\scriptscriptstyle\pm}1.70$ & $54.11{\scriptscriptstyle\pm}1.16$ & $39.92{\scriptscriptstyle\pm}0.40$ & $35.23{\scriptscriptstyle\pm}0.81$ \\
			& 0.3 & $67.74{\scriptscriptstyle\pm}6.55$ & $56.61{\scriptscriptstyle\pm}5.48$ & $63.74{\scriptscriptstyle\pm}8.36$ & $66.55{\scriptscriptstyle\pm}4.78$ & $78.27{\scriptscriptstyle\pm}5.99$ & $70.00{\scriptscriptstyle\pm}1.99$ & $64.91{\scriptscriptstyle\pm}3.77$ & $48.07{\scriptscriptstyle\pm}1.46$ & $39.15{\scriptscriptstyle\pm}0.65$ & $36.08{\scriptscriptstyle\pm}1.07$ \\
			& 0.4 & $59.60{\scriptscriptstyle\pm}6.63$ & $49.36{\scriptscriptstyle\pm}7.67$ & $56.65{\scriptscriptstyle\pm}10.43$ & $42.49{\scriptscriptstyle\pm}9.20$ & $65.98{\scriptscriptstyle\pm}7.74$ & $61.97{\scriptscriptstyle\pm}8.76$ & $52.84{\scriptscriptstyle\pm}4.20$ & $40.30{\scriptscriptstyle\pm}2.43$ & $38.88{\scriptscriptstyle\pm}0.92$ & $31.30{\scriptscriptstyle\pm}1.26$ \\
			& 0.5 & $45.49{\scriptscriptstyle\pm}7.99$ & $36.34{\scriptscriptstyle\pm}4.43$ & $45.96{\scriptscriptstyle\pm}11.46$ & $37.71{\scriptscriptstyle\pm}6.93$ & $43.39{\scriptscriptstyle\pm}17.22$ & $45.23{\scriptscriptstyle\pm}4.31$ & $39.68{\scriptscriptstyle\pm}9.04$ & $31.05{\scriptscriptstyle\pm}2.46$ & $38.33{\scriptscriptstyle\pm}1.18$ & $25.05{\scriptscriptstyle\pm}0.98$ \\
			Random & 0.1 & $79.04{\scriptscriptstyle\pm}1.33$ & $71.60{\scriptscriptstyle\pm}1.50$ & $76.31{\scriptscriptstyle\pm}2.35$ & $80.44{\scriptscriptstyle\pm}1.26$ & $90.14{\scriptscriptstyle\pm}1.36$ & $76.31{\scriptscriptstyle\pm}1.92$ & $73.43{\scriptscriptstyle\pm}1.13$ & $57.47{\scriptscriptstyle\pm}1.07$ & $39.90{\scriptscriptstyle\pm}0.26$ & $40.30{\scriptscriptstyle\pm}0.59$ \\
			& 0.2 & $75.78{\scriptscriptstyle\pm}1.96$ & $65.86{\scriptscriptstyle\pm}4.67$ & $69.92{\scriptscriptstyle\pm}2.64$ & $76.81{\scriptscriptstyle\pm}4.83$ & $87.49{\scriptscriptstyle\pm}2.46$ & $74.82{\scriptscriptstyle\pm}1.80$ & $71.81{\scriptscriptstyle\pm}1.26$ & $53.87{\scriptscriptstyle\pm}1.10$ & $39.46{\scriptscriptstyle\pm}0.49$ & $38.83{\scriptscriptstyle\pm}0.63$ \\
			& 0.3 & $70.89{\scriptscriptstyle\pm}4.45$ & $63.62{\scriptscriptstyle\pm}4.97$ & $65.51{\scriptscriptstyle\pm}5.65$ & $74.46{\scriptscriptstyle\pm}2.04$ & $85.06{\scriptscriptstyle\pm}3.76$ & $69.89{\scriptscriptstyle\pm}2.51$ & $69.25{\scriptscriptstyle\pm}2.02$ & $51.38{\scriptscriptstyle\pm}1.74$ & $37.36{\scriptscriptstyle\pm}1.63$ & $34.16{\scriptscriptstyle\pm}0.89$ \\
			& 0.4 & $65.01{\scriptscriptstyle\pm}4.61$ & $56.95{\scriptscriptstyle\pm}6.46$ & $59.55{\scriptscriptstyle\pm}7.82$ & $69.93{\scriptscriptstyle\pm}3.57$ & $79.66{\scriptscriptstyle\pm}4.48$ & $65.29{\scriptscriptstyle\pm}2.87$ & $66.00{\scriptscriptstyle\pm}3.61$ & $46.84{\scriptscriptstyle\pm}2.39$ & $36.22{\scriptscriptstyle\pm}2.90$ & $34.90{\scriptscriptstyle\pm}0.76$ \\
			& 0.5 & $54.70{\scriptscriptstyle\pm}4.98$ & $43.73{\scriptscriptstyle\pm}5.35$ & $47.62{\scriptscriptstyle\pm}7.76$ & $64.90{\scriptscriptstyle\pm}4.41$ & $72.18{\scriptscriptstyle\pm}6.09$ & $61.40{\scriptscriptstyle\pm}3.09$ & $58.40{\scriptscriptstyle\pm}3.76$ & $42.23{\scriptscriptstyle\pm}1.94$ & $34.15{\scriptscriptstyle\pm}3.32$ & $31.33{\scriptscriptstyle\pm}0.64$ \\
			\midrule
			\textbf{Average} &  & $68.73$ & $60.47$ & $64.03$ & $69.82$ & $80.45$ & $68.87$ & $66.20$ & $49.77$ & $38.54$ & $35.81$ \\
			\bottomrule
		\end{tabular}
	}
\end{table*}

\begin{table*}
	\centering
	\small	
	\caption{Accuracy gain (percentage points) of the proposed method in the NoisyGL Benchmark for all datasets, noise types and noise rates, compared to the GCN baseline. Green and red cells indicate gains and losses greater than 0.5 percentage points in magnitude, respectively.}	
	\label{tab:BenchmarkGainResults}
	\resizebox{\textwidth}{!}{%
\begin{tabular}{llccccccccccc}
	\toprule
	\textbf{Noise} & \textbf{Rate} & \textbf{Cora} & \textbf{CiteSeer} & \textbf{PubMed} & \textbf{Amazon-C} & \textbf{Amazon-P} & \textbf{DBLP} & \textbf{BlogCatalog} & \textbf{Flickr} & \textbf{Amz-Rat.} & \textbf{Roman-Emp.} & \textbf{Average} \\
	\midrule
	Clean & Clean & \cellcolor[HTML]{E2EFDA}+1.37 & \cellcolor[HTML]{E2EFDA}+3.52 & +0.31 & \cellcolor[HTML]{FFC7CE}-2.57 & \cellcolor[HTML]{FFC7CE}-0.56 & \cellcolor[HTML]{E2EFDA}+1.68 & \cellcolor[HTML]{FFC7CE}-1.65 & \cellcolor[HTML]{E2EFDA}+3.00 & \cellcolor[HTML]{E2EFDA}+0.58 & \cellcolor[HTML]{E2EFDA}+5.66 & \cellcolor[HTML]{E2EFDA}+1.13 \\
	Uniform & 0.1 & \cellcolor[HTML]{E2EFDA}+2.25 & \cellcolor[HTML]{E2EFDA}+4.91 & -0.43 & \cellcolor[HTML]{FFC7CE}-1.68 & -0.40 & \cellcolor[HTML]{E2EFDA}+1.01 & \cellcolor[HTML]{FFC7CE}-1.17 & \cellcolor[HTML]{E2EFDA}+1.16 & \cellcolor[HTML]{E2EFDA}+0.59 & \cellcolor[HTML]{E2EFDA}+5.78 & \cellcolor[HTML]{E2EFDA}+1.20 \\
	& 0.2 & \cellcolor[HTML]{E2EFDA}+2.62 & \cellcolor[HTML]{E2EFDA}+6.53 & \cellcolor[HTML]{E2EFDA}+0.76 & \cellcolor[HTML]{FFC7CE}-2.66 & \cellcolor[HTML]{FFC7CE}-1.02 & \cellcolor[HTML]{E2EFDA}+1.38 & +0.23 & \cellcolor[HTML]{E2EFDA}+1.58 & \cellcolor[HTML]{E2EFDA}+1.05 & \cellcolor[HTML]{E2EFDA}+4.80 & \cellcolor[HTML]{E2EFDA}+1.53 \\
	& 0.3 & \cellcolor[HTML]{E2EFDA}+1.02 & \cellcolor[HTML]{E2EFDA}+8.88 & \cellcolor[HTML]{FFC7CE}-3.41 & \cellcolor[HTML]{FFC7CE}-3.06 & \cellcolor[HTML]{E2EFDA}+1.11 & \cellcolor[HTML]{FFC7CE}-0.84 & \cellcolor[HTML]{E2EFDA}+1.00 & \cellcolor[HTML]{E2EFDA}+1.94 & \cellcolor[HTML]{E2EFDA}+1.64 & \cellcolor[HTML]{E2EFDA}+4.08 & \cellcolor[HTML]{E2EFDA}+1.24 \\
	& 0.4 & \cellcolor[HTML]{FFC7CE}-4.51 & \cellcolor[HTML]{E2EFDA}+11.67 & \cellcolor[HTML]{FFC7CE}-1.30 & \cellcolor[HTML]{FFC7CE}-4.80 & \cellcolor[HTML]{E2EFDA}+3.83 & \cellcolor[HTML]{E2EFDA}+2.41 & \cellcolor[HTML]{E2EFDA}+3.38 & \cellcolor[HTML]{E2EFDA}+0.57 & \cellcolor[HTML]{E2EFDA}+0.83 & \cellcolor[HTML]{E2EFDA}+2.84 & \cellcolor[HTML]{E2EFDA}+1.49 \\
	& 0.5 & \cellcolor[HTML]{E2EFDA}+3.54 & \cellcolor[HTML]{E2EFDA}+8.47 & \cellcolor[HTML]{FFC7CE}-4.73 & \cellcolor[HTML]{FFC7CE}-5.47 & \cellcolor[HTML]{E2EFDA}+2.50 & \cellcolor[HTML]{E2EFDA}+5.70 & \cellcolor[HTML]{E2EFDA}+1.08 & \cellcolor[HTML]{E2EFDA}+0.56 & -0.10 & \cellcolor[HTML]{E2EFDA}+6.07 & \cellcolor[HTML]{E2EFDA}+1.76 \\
	Pair & 0.1 & \cellcolor[HTML]{E2EFDA}+3.15 & \cellcolor[HTML]{E2EFDA}+4.57 & \cellcolor[HTML]{E2EFDA}+1.38 & \cellcolor[HTML]{FFC7CE}-2.45 & \cellcolor[HTML]{FFC7CE}-2.10 & \cellcolor[HTML]{E2EFDA}+2.74 & -0.06 & \cellcolor[HTML]{E2EFDA}+2.52 & \cellcolor[HTML]{E2EFDA}+0.85 & \cellcolor[HTML]{E2EFDA}+5.51 & \cellcolor[HTML]{E2EFDA}+1.61 \\
	& 0.2 & \cellcolor[HTML]{FFC7CE}-6.03 & \cellcolor[HTML]{E2EFDA}+8.18 & \cellcolor[HTML]{E2EFDA}+0.95 & \cellcolor[HTML]{FFC7CE}-0.70 & \cellcolor[HTML]{FFC7CE}-1.67 & \cellcolor[HTML]{E2EFDA}+2.94 & \cellcolor[HTML]{E2EFDA}+3.75 & \cellcolor[HTML]{E2EFDA}+2.56 & \cellcolor[HTML]{E2EFDA}+0.85 & \cellcolor[HTML]{E2EFDA}+2.32 & \cellcolor[HTML]{E2EFDA}+1.32 \\
	& 0.3 & \cellcolor[HTML]{E2EFDA}+2.38 & \cellcolor[HTML]{E2EFDA}+2.95 & \cellcolor[HTML]{E2EFDA}+0.83 & \cellcolor[HTML]{FFC7CE}-4.40 & \cellcolor[HTML]{FFC7CE}-0.99 & \cellcolor[HTML]{E2EFDA}+7.44 & \cellcolor[HTML]{E2EFDA}+4.22 & \cellcolor[HTML]{E2EFDA}+2.39 & \cellcolor[HTML]{E2EFDA}+0.74 & \cellcolor[HTML]{E2EFDA}+5.77 & \cellcolor[HTML]{E2EFDA}+2.13 \\
	& 0.4 & \cellcolor[HTML]{E2EFDA}+5.58 & \cellcolor[HTML]{E2EFDA}+5.89 & \cellcolor[HTML]{E2EFDA}+0.98 & \cellcolor[HTML]{FFC7CE}-18.43 & \cellcolor[HTML]{E2EFDA}+1.08 & \cellcolor[HTML]{E2EFDA}+9.81 & \cellcolor[HTML]{E2EFDA}+6.09 & \cellcolor[HTML]{E2EFDA}+1.48 & \cellcolor[HTML]{E2EFDA}+2.15 & \cellcolor[HTML]{E2EFDA}+4.94 & \cellcolor[HTML]{E2EFDA}+1.96 \\
	& 0.5 & \cellcolor[HTML]{E2EFDA}+1.34 & \cellcolor[HTML]{E2EFDA}+0.86 & \cellcolor[HTML]{E2EFDA}+2.97 & \cellcolor[HTML]{FFC7CE}-1.52 & \cellcolor[HTML]{FFC7CE}-1.57 & \cellcolor[HTML]{E2EFDA}+5.24 & \cellcolor[HTML]{E2EFDA}+4.32 & \cellcolor[HTML]{E2EFDA}+2.14 & \cellcolor[HTML]{E2EFDA}+3.92 & \cellcolor[HTML]{E2EFDA}+3.35 & \cellcolor[HTML]{E2EFDA}+2.11 \\
	Random & 0.1 & \cellcolor[HTML]{E2EFDA}+0.85 & \cellcolor[HTML]{E2EFDA}+5.59 & \cellcolor[HTML]{E2EFDA}+2.52 & \cellcolor[HTML]{FFC7CE}-2.46 & \cellcolor[HTML]{E2EFDA}+2.09 & \cellcolor[HTML]{E2EFDA}+0.91 & \cellcolor[HTML]{E2EFDA}+0.80 & \cellcolor[HTML]{E2EFDA}+2.30 & \cellcolor[HTML]{E2EFDA}+1.05 & \cellcolor[HTML]{E2EFDA}+4.71 & \cellcolor[HTML]{E2EFDA}+1.84 \\
	& 0.2 & \cellcolor[HTML]{E2EFDA}+1.51 & \cellcolor[HTML]{E2EFDA}+4.75 & \cellcolor[HTML]{FFC7CE}-2.57 & \cellcolor[HTML]{FFC7CE}-3.38 & \cellcolor[HTML]{E2EFDA}+0.70 & \cellcolor[HTML]{E2EFDA}+2.32 & \cellcolor[HTML]{E2EFDA}+1.11 & +0.15 & \cellcolor[HTML]{E2EFDA}+1.15 & \cellcolor[HTML]{E2EFDA}+4.89 & \cellcolor[HTML]{E2EFDA}+1.06 \\
	& 0.3 & \cellcolor[HTML]{E2EFDA}+1.17 & \cellcolor[HTML]{E2EFDA}+7.18 & \cellcolor[HTML]{FFC7CE}-1.01 & \cellcolor[HTML]{FFC7CE}-1.33 & \cellcolor[HTML]{E2EFDA}+2.82 & \cellcolor[HTML]{E2EFDA}+3.29 & \cellcolor[HTML]{E2EFDA}+3.44 & \cellcolor[HTML]{E2EFDA}+1.81 & -0.33 & \cellcolor[HTML]{E2EFDA}+3.07 & \cellcolor[HTML]{E2EFDA}+2.01 \\
	& 0.4 & \cellcolor[HTML]{E2EFDA}+2.42 & \cellcolor[HTML]{E2EFDA}+9.15 & \cellcolor[HTML]{E2EFDA}+2.57 & \cellcolor[HTML]{FFC7CE}-2.94 & \cellcolor[HTML]{E2EFDA}+2.72 & \cellcolor[HTML]{E2EFDA}+2.53 & \cellcolor[HTML]{E2EFDA}+4.25 & -0.32 & +0.04 & \cellcolor[HTML]{E2EFDA}+4.27 & \cellcolor[HTML]{E2EFDA}+2.47 \\
	& 0.5 & \cellcolor[HTML]{FFC7CE}-0.89 & \cellcolor[HTML]{E2EFDA}+1.97 & \cellcolor[HTML]{E2EFDA}+1.38 & +0.17 & \cellcolor[HTML]{E2EFDA}+2.33 & \cellcolor[HTML]{E2EFDA}+7.14 & \cellcolor[HTML]{E2EFDA}+0.79 & \cellcolor[HTML]{E2EFDA}+1.57 & \cellcolor[HTML]{E2EFDA}+1.54 & \cellcolor[HTML]{E2EFDA}+3.11 & \cellcolor[HTML]{E2EFDA}+1.91 \\
	\midrule
	\textbf{Average} &  & \cellcolor[HTML]{E2EFDA}+1.11 & \cellcolor[HTML]{E2EFDA}+5.94 & +0.07 & \cellcolor[HTML]{FFC7CE}-3.61 & \cellcolor[HTML]{E2EFDA}+0.68 & \cellcolor[HTML]{E2EFDA}+3.48 & \cellcolor[HTML]{E2EFDA}+1.97 & \cellcolor[HTML]{E2EFDA}+1.59 & \cellcolor[HTML]{E2EFDA}+1.03 & \cellcolor[HTML]{E2EFDA}+4.45 & \cellcolor[HTML]{E2EFDA}+1.67 \\
	\bottomrule
\end{tabular}
	}
\end{table*}

Table~\ref{tab:AverageAccuracybyMethod} shows the average accuracy obtained by each method in each dataset, considering all scenarios: clean labels and all types of noise and noise rates. The proposed method (PCC+GCN) achieved the best results in four of the ten datasets: CiteSeer, Amazon-Photos, Flickr, and Amazon-Ratings. It also achieved the highest overall average accuracy across all datasets ($60.27\%$), followed by NRGNN ($59.31\%$), which achieved the best result in three datasets. NRGNN was the only other method to outperform the GCN baseline ($58.60\%$), achieving an overall average accuracy of $59.31\%$.

\begin{table*}[htbp]
	\centering
	\small
	\caption{Global average accuracy by method (clean + all noisy types and rates). \textit{---} indicates that the method does not report results for the corresponding dataset; averages involving RNCGLN are therefore computed over the available datasets.}	
	\label{tab:AverageAccuracybyMethod}
	\resizebox{\textwidth}{!}{%
		\begin{tabular}{lccccccccccc}
			\toprule
			\textbf{Method} & \textbf{Cora} & \textbf{CiteSeer} & \textbf{PubMed} & \textbf{Amazon-C} & \textbf{Amazon-P} & \textbf{DBLP} & \textbf{BlogCatalog} & \textbf{Flickr} & \textbf{Amz-Rat.} & \textbf{Roman-Emp.} & \textbf{Average} \\
			\midrule
			GCN & 67.62 & 54.52 & 63.96 & 73.42 & 79.78 & 65.39 & 64.22 & 48.18 & 37.50 & 31.36 & 58.60 \\
			CGNN & 66.37 & 51.15 & 56.36 & 43.74 & 47.19 & 58.51 & 24.15 & 11.77 & 34.09 & 18.83 & 41.21 \\
			CLNode & 66.12 & 53.88 & 62.66 & \cellcolor[HTML]{A8D08D}\textbf{74.70} & 79.67 & 63.77 & 63.35 & 46.64 & 36.79 & 30.67 & 57.82 \\
			CP & 68.22 & 54.99 & 64.36 & 72.94 & 78.74 & 66.79 & 63.25 & 42.75 & 37.62 & 30.47 & 58.01 \\
			CR-GNN & 67.87 & 53.49 & 64.84 & 45.74 & 36.09 & 65.97 & 59.48 & 32.14 & 35.10 & 23.37 & 48.41 \\
			DGNN & 57.72 & 47.97 & 61.64 & 49.56 & 55.05 & 62.85 & 42.55 & 17.17 & 34.34 & 21.24 & 45.01 \\
			NRGNN & \cellcolor[HTML]{A8D08D}\textbf{71.00} & 60.10 & 60.05 & 65.91 & 71.40 & \cellcolor[HTML]{A8D08D}\textbf{71.81} & \cellcolor[HTML]{A8D08D}\textbf{70.09} & 41.02 & 36.34 & 45.36 & 59.31 \\
			PIGNN & 66.44 & 57.54 & \cellcolor[HTML]{A8D08D}\textbf{66.01} & 74.18 & 79.75 & 69.15 & 52.88 & 47.77 & 37.22 & 30.21 & 58.11 \\
			RNCGLN & 70.93 & 59.70 & --- & 61.58 & 68.48 & 63.28 & 55.76 & 24.08 & 31.17 & \cellcolor[HTML]{A8D08D}\textbf{47.94} & 53.66 \\
			RTGNN & 64.01 & 50.72 & 63.10 & 62.06 & 76.43 & 61.70 & 69.73 & 39.08 & 36.19 & 45.66 & 56.87 \\
			UnionNET & 68.46 & 58.41 & 63.79 & 33.47 & 29.30 & 65.83 & 49.61 & 21.07 & 35.32 & 15.21 & 44.05 \\
			PCC+GCN & 68.73 & \cellcolor[HTML]{A8D08D}\textbf{60.47} & 64.03 & 69.82 & \cellcolor[HTML]{A8D08D}\textbf{80.45} & 68.87 & 66.20 & \cellcolor[HTML]{A8D08D}\textbf{49.77} & \cellcolor[HTML]{A8D08D}\textbf{38.54} & 35.81 & \cellcolor[HTML]{A8D08D}\textbf{60.27} \\
			\bottomrule
		\end{tabular}
	}
\end{table*}

Table~\ref{tab:AverageAccuracyRanking} shows the accuracy ranking of the methods for each dataset. The proposed method achieved the best overall position, with an average rank of $2.60$, followed by NRGNN ($4.30$), while GCN and PIGNN tied for third place ($4.50$), and CP ranked fifth ($4.70$). More importantly, PCC+GCN was never ranked lower than fifth among the evaluated methods in any of the ten datasets. It ranked first in four datasets and among the top three in seven datasets, indicating a consistently strong performance across datasets rather than being driven by a small number of particularly favorable cases.

\begin{table*}[htbp]
	\centering
	\small
	\caption{Ranking of methods by average accuracy in each dataset (clean + all noisy types and rates). \textit{---} indicates that the method does not report results for the corresponding dataset; averages involving RNCGLN are therefore computed over the available datasets.}	
	\label{tab:AverageAccuracyRanking}
	\resizebox{\textwidth}{!}{%
		\begin{tabular}{lccccccccccc}
			\toprule
			\textbf{Method} & \textbf{Cora} & \textbf{CiteSeer} & \textbf{PubMed} & \textbf{Amazon-C} & \textbf{Amazon-P} & \textbf{DBLP} & \textbf{BlogCatalog} & \textbf{Flickr} & \textbf{Amz-Rat.} & \textbf{Roman-Emp.} & \textbf{Average Rank} \\
			\midrule
			GCN & 7 & 7 & 5 & 3 & 2 & 7 & 4 & 2 & 3 & 5 & 4.50 \\
			CGNN & 9 & 10 & 11 & 11 & 10 & 12 & 12 & 12 & 11 & 11 & 10.90 \\
			CLNode & 10 & 8 & 8 & \cellcolor[HTML]{A8D08D}\textbf{1} & 4 & 8 & 5 & 4 & 5 & 6 & 5.90 \\
			CP & 5 & 6 & 3 & 4 & 5 & 4 & 6 & 5 & 2 & 7 & 4.70 \\
			CR-GNN & 6 & 9 & 2 & 10 & 11 & 5 & 7 & 8 & 9 & 9 & 7.60 \\
			DGNN & 12 & 12 & 9 & 9 & 9 & 10 & 11 & 11 & 10 & 10 & 10.30 \\
			NRGNN & \cellcolor[HTML]{A8D08D}\textbf{1} & 2 & 10 & 6 & 7 & \cellcolor[HTML]{A8D08D}\textbf{1} & \cellcolor[HTML]{A8D08D}\textbf{1} & 6 & 6 & 3 & 4.30 \\
			PIGNN & 8 & 5 & \cellcolor[HTML]{A8D08D}\textbf{1} & 2 & 3 & 2 & 9 & 3 & 4 & 8 & 4.50 \\
			RNCGLN & 2 & 3 & --- & 8 & 8 & 9 & 8 & 9 & 12 & \cellcolor[HTML]{A8D08D}\textbf{1} & 6.67 \\
			RTGNN & 11 & 11 & 7 & 7 & 6 & 11 & 2 & 7 & 7 & 2 & 7.10 \\
			UnionNET & 4 & 4 & 6 & 12 & 12 & 6 & 10 & 10 & 8 & 12 & 8.40 \\
			PCC+GCN & 3 & \cellcolor[HTML]{A8D08D}\textbf{1} & 4 & 5 & \cellcolor[HTML]{A8D08D}\textbf{1} & 3 & 3 & \cellcolor[HTML]{A8D08D}\textbf{1} & \cellcolor[HTML]{A8D08D}\textbf{1} & 4 & \cellcolor[HTML]{A8D08D}\textbf{2.60} \\
			\bottomrule
		\end{tabular}
		}
\end{table*}

\begin{figure*}
	\centering
	\includegraphics[width=1\linewidth]{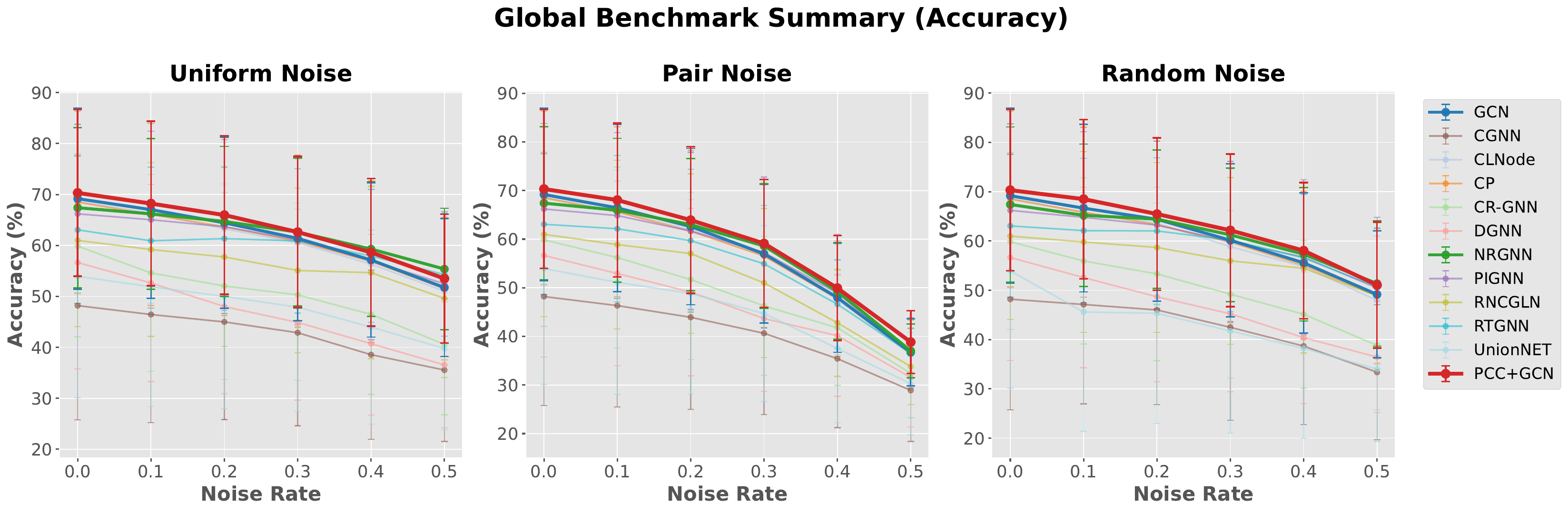}
	\caption{Average classification accuracy of all benchmark methods across all datasets for different label noise types and rates.}
	\label{fig:SummaryClassicAccuracy}
\end{figure*}

Figure~\ref{fig:SummaryClassicAccuracy} shows the average classification accuracy of all benchmark methods across all datasets for each label noise type and rate. PCC+GCN achieved the highest average accuracy in all Pair-noise scenarios. For Uniform noise, it achieved the best results at noise rates up to $30\%$, while for Random noise, it remained the best up to $40\%$. Overall, PCC+GCN was the best-performing method in $12$ of the $15$ noisy-label scenarios. In the three remaining scenarios, its average accuracy remained close to that of the best-performing method, NRGNN.

\begin{figure*}
	\centering
	\includegraphics[width=1\linewidth]{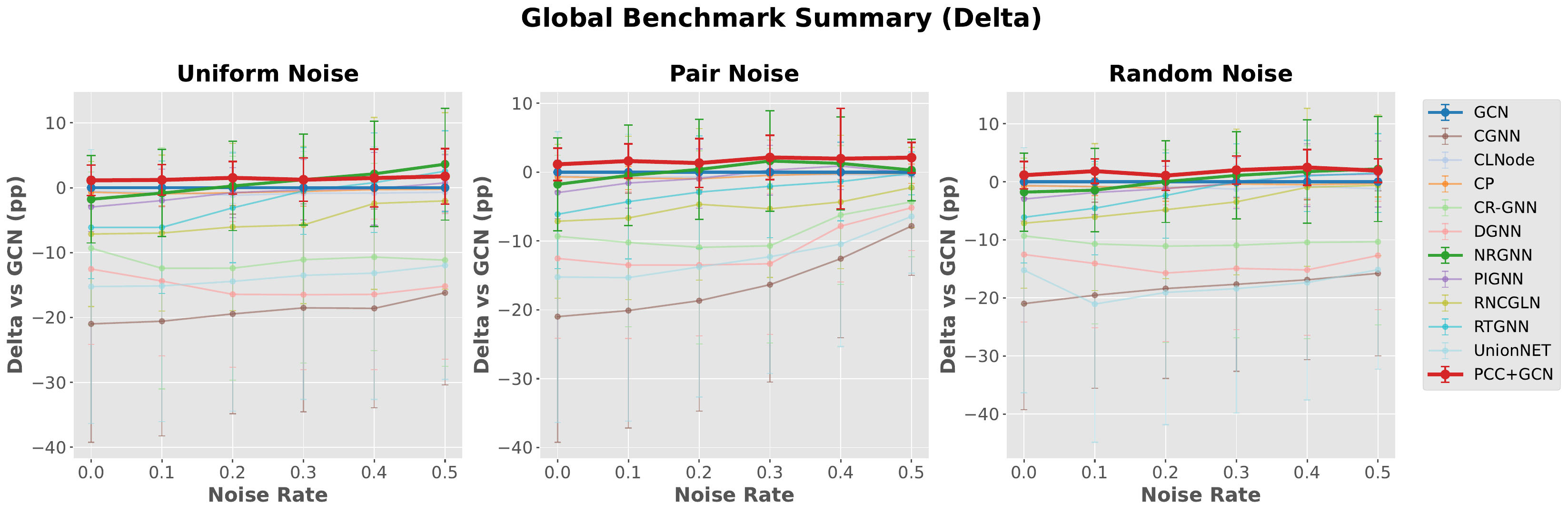}
	\caption{Average improvement in classification accuracy of all benchmark methods across all datasets relative to the baseline GCN, for different label noise types and rates.}
	\label{fig:DeltaClassicAccuracy}
\end{figure*}

Figure~\ref{fig:DeltaClassicAccuracy} shows the average accuracy improvement of all benchmark methods across all datasets for each label noise type and rate. PCC+GCN was the only method that improved over the GCN baseline in all tested noisy-label scenarios. Considering the clean setting and all five noise rates, its average improvements were $1.39$, $1.71$, and $1.74$ percentage points for Uniform, Pair, and Random noise, respectively.

Taken together, these results indicate that PCC+GCN combines strong average performance with consistent robustness across datasets and noise conditions. Although it does not achieve the highest accuracy in every individual dataset or noise scenario, it obtains both the highest overall average accuracy and the best average rank among the evaluated methods, while remaining the only method to improve over the GCN baseline across all tested noisy-label scenarios.

\section{Instance-Dependent Noise} \label{sec:InstanceNoise}

This section compares the proposed PCC+GCN method with the three best-performing competing methods in the NoisyGL benchmark according to the overall average accuracy reported in the previous section, namely CP, NRGNN, and PIGNN, as well as the baseline GCN. The comparison is conducted across the ten datasets described in Table~\ref{tab:Datasets} under instance-dependent label noise, with noise rates ranging from $10\%$ to $50\%$. In addition, the clean setting is included as a reference without label corruption.

\begin{figure}
	\centering
	\includegraphics[width=1\linewidth]{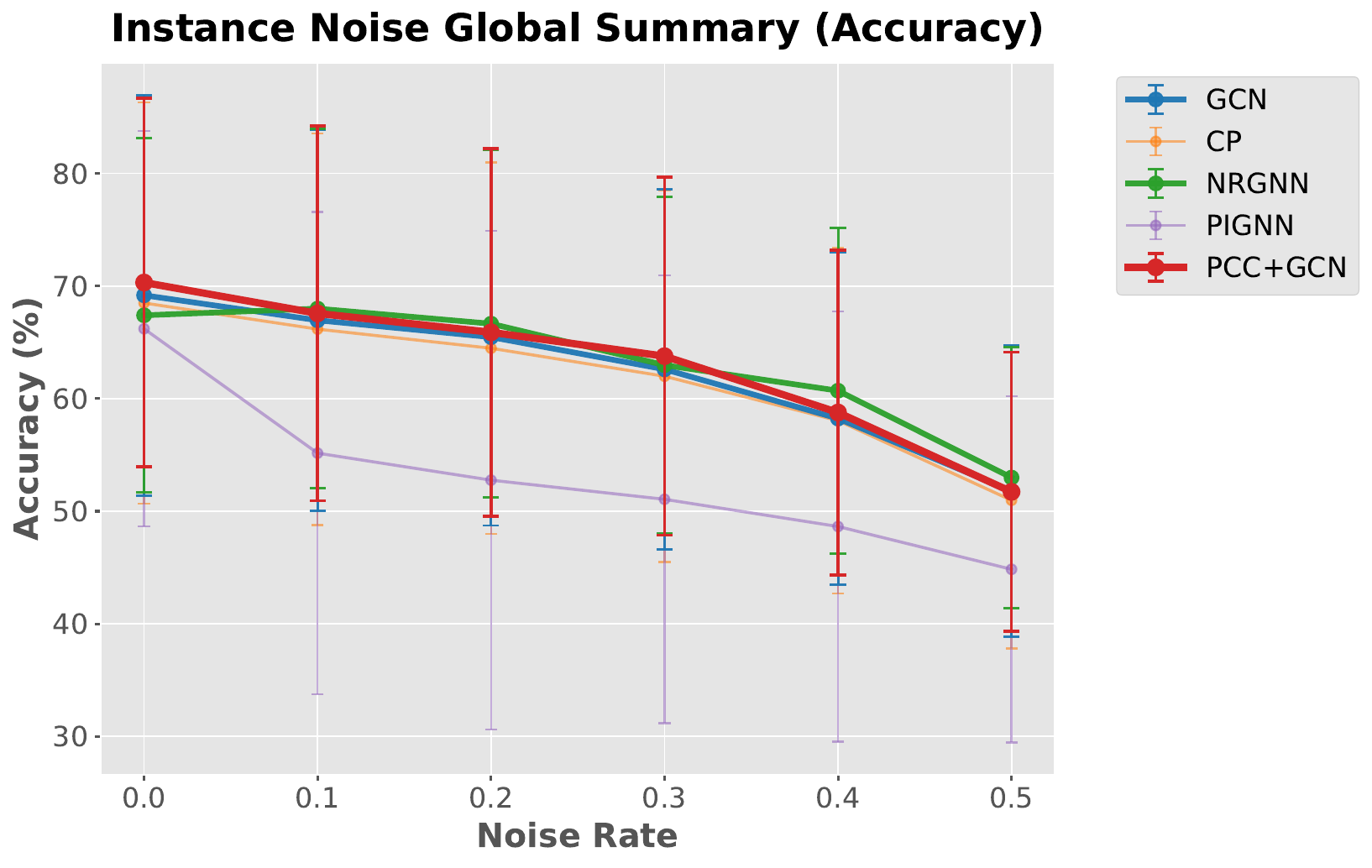}
	\caption{Average classification accuracy of GCN, CP, NRGNN, PIGNN, and PCC+GCN across all datasets under instance-dependent label noise at different noise rates.}
	\label{fig:SummaryInstanceAccuracy}
\end{figure}

Figure~\ref{fig:SummaryInstanceAccuracy} shows the average classification accuracy of the methods across all datasets for each instance-dependent noise rate. NRGNN achieved the best results at noise rates of $10\%$, $20\%$, $40\%$, and $50\%$, while PCC+GCN achieved the best results in the clean setting and at the $30\%$ noise rate. Overall, NRGNN achieved the highest average accuracy ($63.12\%$), closely followed by PCC+GCN ($63.00\%$). The baseline GCN achieved $62.37\%$, while CP and PIGNN obtained lower average accuracies than the baseline.

\begin{figure}
	\centering
	\includegraphics[width=1\linewidth]{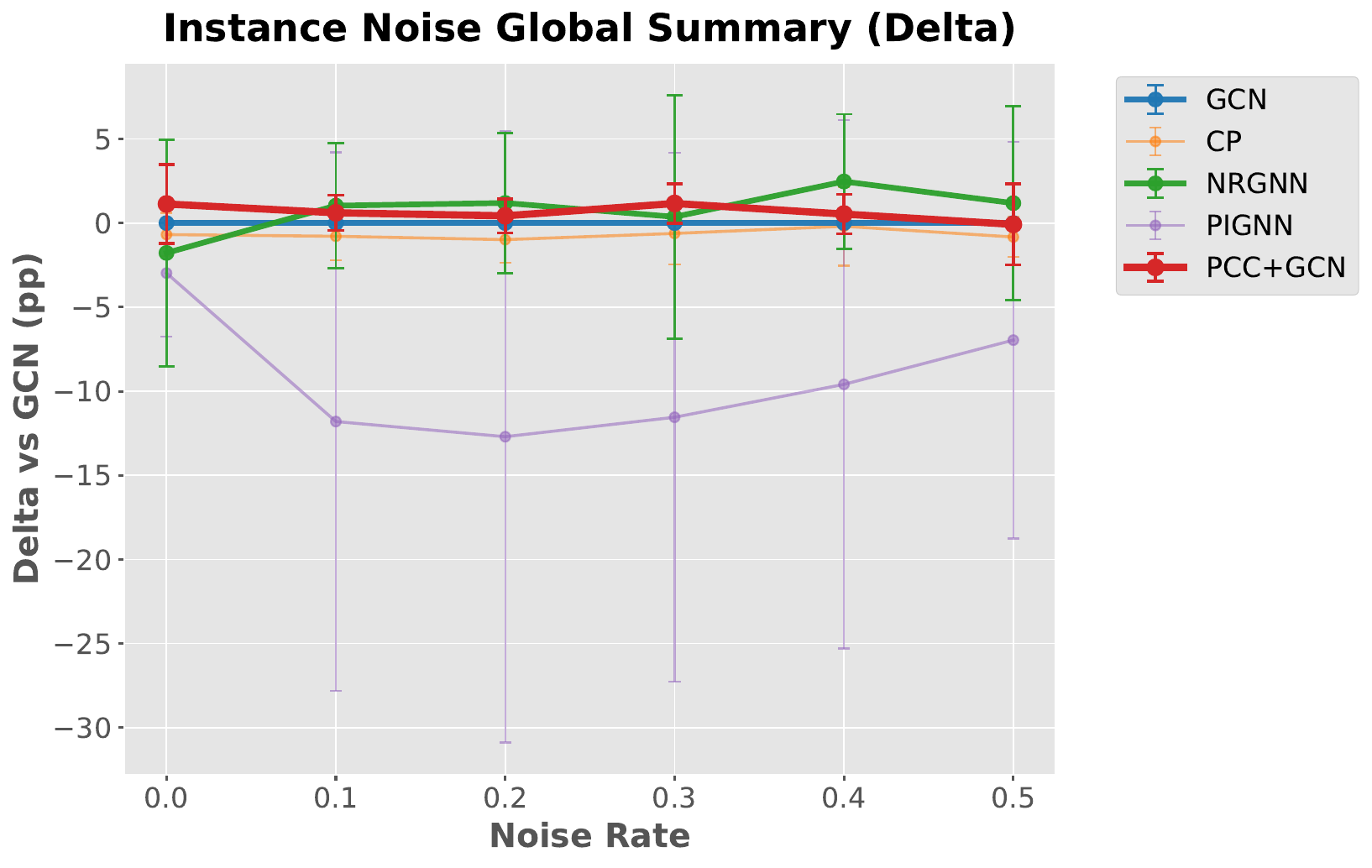}
	\caption{Average improvement in classification accuracy of CP, NRGNN, PIGNN, and PCC+GCN across all datasets relative to the baseline GCN under instance-dependent label noise at different noise rates.}
	\label{fig:SummaryInstanceDelta}
\end{figure}

Figure~\ref{fig:SummaryInstanceDelta} shows the average accuracy improvement of the methods relative to the baseline GCN across all datasets for each instance-dependent noise rate. PCC+GCN improved the average accuracy relative to the baseline in the clean setting and at all instance-dependent noise rates except $50\%$, where its average accuracy was $0.08$ percentage points below the baseline. In contrast, NRGNN improved the accuracy at all instance-dependent noise rates but decreased it in the clean setting. Overall, NRGNN and PCC+GCN improved the average accuracy relative to the baseline GCN by $0.75$ and $0.63$ percentage points, respectively. CP and PIGNN decreased the average accuracy relative to the baseline in all evaluated settings.

\begin{figure*}
	\centering
	\includegraphics[width=1\linewidth]{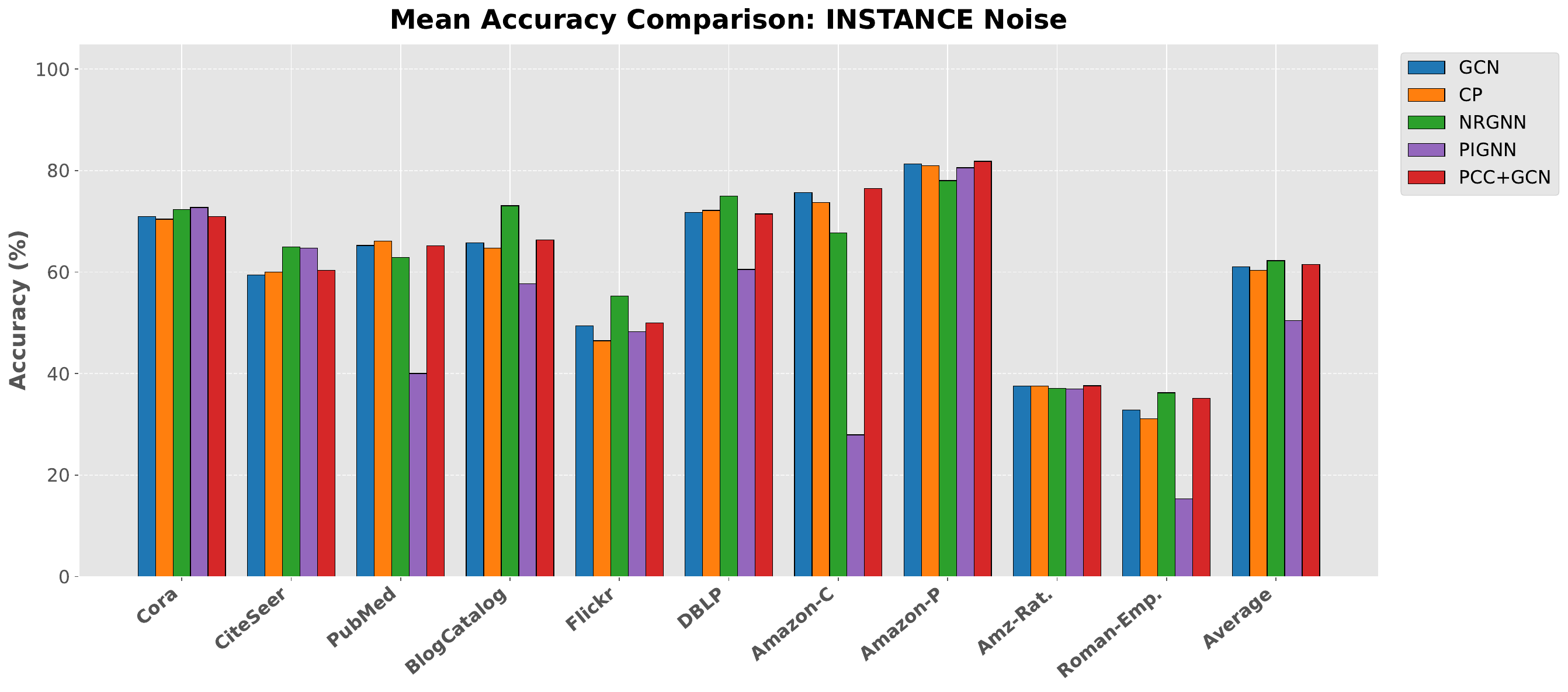}
	\caption{Average classification accuracy of GCN, CP, NRGNN, PIGNN, and PCC+GCN on each dataset under instance-dependent label noise, considering noise rates from $10\%$ to $50\%$.}
	\label{fig:BarAvgInstance}
\end{figure*}

Figure~\ref{fig:BarAvgInstance} shows the average accuracy of each method on each dataset, considering the instance-dependent noise rates from $10\%$ to $50\%$. NRGNN achieved the best results on five datasets, followed by PCC+GCN, which achieved the best results on three datasets. CP and PIGNN achieved the best result on one dataset each. On average across the ten datasets and five noise rates, only NRGNN ($62.26\%$) and PCC+GCN ($61.54\%$) outperformed the baseline GCN ($61.01\%$).

\begin{figure*}
	\centering
	\includegraphics[width=1\linewidth]{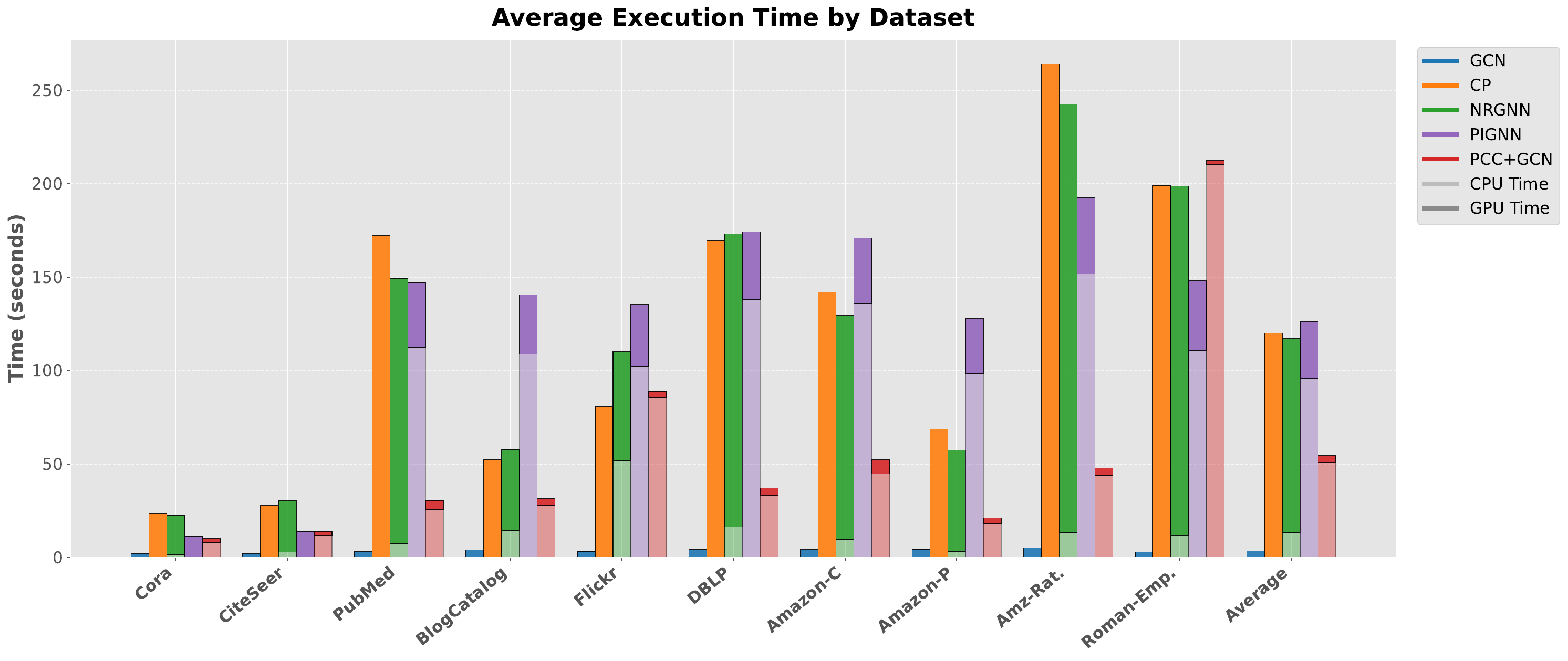}
	\caption{Average CPU and GPU execution times of GCN, CP, NRGNN, PIGNN, and PCC+GCN on each dataset under instance-dependent label noise. The stacked bars represent the total execution time (CPU + GPU).}
	\label{fig:StackedTimes}
\end{figure*}

Figure~\ref{fig:StackedTimes} shows the average execution time of each method on each dataset, with CPU and GPU times shown separately and stacked to represent the total execution time. As expected, the baseline GCN was the fastest method on all datasets because it does not include the additional noise-handling procedures employed by the robust methods. Excluding GCN, PCC+GCN was the fastest method on eight of the ten datasets, while CP and PIGNN were the fastest on one dataset each. On average, PCC+GCN required $54.52$ seconds per run, less than half the $117.11$ seconds required by NRGNN, the second-fastest robust method. The distribution of execution time also differed among the methods: most of the execution time of CP and NRGNN was spent on GPU computation, whereas PIGNN and PCC+GCN spent most of their execution time on CPU computation.

Taken together, these results show that PCC+GCN remains competitive under instance-dependent label noise. Although NRGNN achieved the highest overall average accuracy, PCC+GCN followed closely and outperformed the baseline GCN on average, while achieving the best results on three of the ten datasets when considering the noisy scenarios. At the same time, PCC+GCN exhibited a substantially lower execution time than the other robust methods, being the fastest among them on eight datasets and requiring less than half the average execution time of NRGNN. These results indicate that PCC+GCN provides a favorable balance between robustness to instance-dependent label noise, classification performance, and computational efficiency.

\section{Discussion} \label{sec:Discussion}

The experimental results indicate that PCC-based label refinement can be an effective preprocessing strategy for improving the robustness of GCNs under label noise. Rather than modifying the GCN architecture or training objective, the proposed framework acts directly on the supervision provided to the classifier. This separation between label refinement and representation learning is one of the main characteristics of PCC+GCN: PCC operates before GCN training, while the GCN itself remains unchanged and is trained on the original graph structure and node features using the refined labels.

The hyperparameter analysis also showed that the behavior of PCC is strongly dataset-dependent. In particular, the greedy-walk probability and distance exponent exhibited markedly different optimal configurations across Cora, CiteSeer, and PubMed. CiteSeer favored purely random walks, whereas Cora and PubMed benefited from a nonzero greedy component. This suggests that the balance between exploration and domination-guided movement depends on the structural and feature characteristics of the graph. In contrast, the label-refinement thresholds showed greater consistency across datasets, with low removal and relabeling thresholds generally providing the best average performance. These results indicate that the movement dynamics of PCC may require greater adaptation to the dataset than the refinement decision itself.

The graph-augmentation experiments further showed that adding feature-based k-NN edges is not universally beneficial. The Same-Label strategy was consistently the most effective augmentation mode, while the more permissive Non-Conflicting and Full modes frequently degraded performance, particularly as the number of added neighbors increased. This behavior suggests that indiscriminate graph densification may propagate unreliable or structurally inconsistent information during PCC refinement. Restricting additional edges to pairs of labeled nodes that agree in class appears to provide a more conservative way of exploiting feature similarity without substantially altering the original graph topology.

Under the conventional NoisyGL noise models, PCC+GCN achieved the highest overall average accuracy and the best average rank among the evaluated methods. Although it was not the best-performing method on every dataset or noise condition, its consistently high ranking across datasets indicates that its performance was not driven by a small number of favorable cases. Moreover, the proposed method improved over the GCN baseline in all tested noisy-label scenarios when results were averaged across datasets. Another relevant pattern is that the benefit of PCC-based refinement tends to become more pronounced at higher noise levels. When averaged across Uniform, Pair, and Random noise, the accuracy gain relative to GCN was generally larger at noise rates between $30\%$ and $50\%$ than at lower corruption levels. Although this trend is not strictly monotonic, it suggests that the refinement stage becomes particularly useful as label corruption becomes more severe.

The results under instance-dependent noise present a more nuanced picture. NRGNN achieved a slightly higher overall average accuracy than PCC+GCN, indicating that the proposed refinement strategy does not dominate more tightly integrated robust graph-learning methods under all corruption processes. Nevertheless, PCC+GCN remained competitive, outperforming the GCN baseline on average and achieving the best results on three datasets. More importantly, it required substantially less execution time than the other robust methods. Its mean runtime was approximately two times lower than those of CP, NRGNN, and PIGNN, and PCC+GCN was the fastest robust method on eight of the ten datasets. This indicates that its computational advantage is broadly distributed across datasets rather than being driven by a single favorable case, and suggests a favorable trade-off between robustness and computational cost.

These findings also highlight some limitations of the proposed approach. PCC+GCN depends on the quality of the graph used during the refinement stage and on the suitability of its hyperparameters for the dataset. The results with Amazon-Computers in the conventional benchmark, where the proposed method produced a negative average gain relative to GCN, show that label refinement is not universally beneficial. In addition, the current study evaluates only node classification and relies on the datasets and noise-generation procedures provided by NoisyGL. Future work could investigate adaptive strategies for selecting PCC hyperparameters, alternative graph-construction mechanisms, and the extension of the framework to other GNN architectures and graph-learning tasks.

\section{Conclusion} \label{sec:Conclusion}

This work proposed PCC+GCN, a hybrid framework that uses Particle Competition and Cooperation as a graph-based label-refinement stage before Graph Convolutional Network training. The objective was to improve robustness to noisy supervision without modifying the underlying GCN architecture or training objective.

The experimental results showed that the proposed approach is effective under a broad range of label-noise conditions. In the conventional NoisyGL benchmark, PCC+GCN achieved the highest overall average accuracy and the best average rank among the evaluated methods, while improving over the GCN baseline across all tested noisy-label scenarios when performance was averaged across datasets. The hyperparameter analysis also showed that the PCC dynamics are dataset-dependent, while conservative graph augmentation based on label agreement was generally more effective than more permissive augmentation strategies.

Under instance-dependent label noise, PCC+GCN remained competitive with the best-performing robust methods. Although NRGNN achieved a slightly higher overall average accuracy, PCC+GCN achieved a higher overall average accuracy than the baseline GCN and substantially lower execution times, being the fastest robust method on most datasets. These results indicate that PCC-based label refinement provides an effective and computationally efficient way to improve GCN robustness under noisy labels.

Overall, the results support the use of PCC as a preprocessing mechanism for graph neural network training, particularly when robustness to label noise is required without introducing substantial additional model complexity. Future work may investigate adaptive selection of PCC hyperparameters, alternative graph-construction strategies, and the extension of the framework to other graph neural network architectures and graph-learning tasks.

\section*{Data and Code Availability}

The source code used for the hyperparameter analysis, together with the
complete experimental results and additional figures, is publicly available at
\url{https://github.com/fbreve/PCC-GCN} and archived at Zenodo
\citep{breve2026pccgcn}.

The implementation used for the benchmark and instance-dependent
label-noise experiments, including the GPU-accelerated implementation of
the instance-dependent label-noise generator, is publicly available at
\url{https://github.com/fbreve/NoisyGL} and archived at Zenodo
\citep{breve2026noisygl}.

\section*{Declaration of generative AI and AI-assisted technologies in the manuscript preparation process}

During the preparation of this work, the author used ChatGPT (OpenAI) and Perplexity to assist with language editing, manuscript organization, and the drafting and revision of portions of the text. The author also used AI-assisted tools, including Perplexity, Google Gemini, Anthropic Claude Sonnet, and OpenAI GPT models, to assist with the development, debugging, and refinement of the research code. All AI-assisted text and code were reviewed, edited, tested, and validated by the author as appropriate. The author takes full responsibility for the content of the published article and for the correctness of the reported implementation and results.




%
%
%





\bibliographystyle{cas-model2-names}

\bibliography{pcc_gcn}

\bio{figs/breve}
Fabricio Breve is a Professor at São Paulo State University (UNESP), Brazil. 
His research interests include machine learning, graph-based learning, 
semi-supervised learning, graph neural networks, computer vision, and 
pattern recognition. He has worked extensively on Particle Competition and 
Cooperation methods and their applications to classification, clustering, 
image processing, and learning under label noise. His current research 
focuses on robust learning methods, hybrid graph-based models, and the 
application of deep learning techniques to complex classification problems.
\endbio

\end{document}